\documentclass{article}

\PassOptionsToPackage{numbers}{natbib}
\usepackage[final]{neurips_2020}

\usepackage[utf8]{inputenc} % allow utf-8 input
\usepackage[T1]{fontenc}    % use 8-bit T1 fonts
\usepackage[breaklinks=true,colorlinks,bookmarks=false]{hyperref}
\usepackage{url}            % simple URL typesetting
\usepackage{booktabs}       % professional-quality tables
\usepackage{amsfonts}       % blackboard math symbols
\usepackage{nicefrac}       % compact symbols for 1/2, etc.
\usepackage{microtype}      % microtypography
\usepackage{amsthm}
\usepackage{amsmath}
\usepackage{graphicx}
\usepackage{amssymb}
\usepackage{diagbox}
\usepackage{multirow}
\usepackage{nicefrac}
\usepackage{bm}
\usepackage{subcaption}
\usepackage{color,xcolor}
\usepackage{caption}
\usepackage{wrapfig}
\usepackage{algorithm, algpseudocode}
\newcommand{\vect}[1]{\boldsymbol{\mathbf{#1}}}

\let\oldReturn\Return
\renewcommand{\Return}{\State\oldReturn}

\newtheorem{lemma}{Lemma}
\newtheorem{theorem}{Theorem}

\newtheorem{remark}{Remark}

\newtheorem{assumption}{Assumption}

\DeclareMathOperator*{\argmin}{arg\,min}
\DeclareMathOperator*{\argmax}{arg\,max}
\title{Adversarial Distributional Training for \\ Robust Deep Learning}

\renewcommand\footnotemark{}
\author{
Yinpeng Dong$^*$, Zhijie Deng$^*$, Tianyu Pang, Hang Su, Jun Zhu$^\dagger$ \thanks{$^*$Equal contribution. $\dagger$  Corresponding author.}\\
  Dept. of Comp. Sci. \& Tech., Institute for AI, BNRist Center\\
Tsinghua-Bosch Joint ML Center, THBI Lab, Tsinghua University, Beijing, 100084 China \\
  \texttt{\scriptsize \{dyp17, dzj17, pty17\}@mails.tsinghua.edu.cn, \{suhangss, dcszj\}@mail.tsinghua.edu.cn}\\
}

\begin{document}

\maketitle
%\vspace{-2ex}
\begin{abstract}
%\junz{the abstract needs revision: the statement on AT (e.g., not perform well enough) is too general, not providing a good (explicit and convincing) motivation for moving from a single worst-case attack to a worst-case distribution over attacks, which is the key of ADT. Likely for the Introduction. For technical contributions, seems entropic regularization is a key to succeed, which should be mentioned; also the parameterization from explicit to implicit looks also interesting, which should be better summarized in Abstract and Intro.}
%\vspace{-1ex}
  Adversarial training (AT) is among the most effective techniques to improve model robustness by augmenting training data with adversarial examples. However, most existing AT methods adopt a specific attack to craft adversarial examples, leading to the unreliable robustness against other unseen attacks.
  Besides, a single attack algorithm could be insufficient to explore the space of perturbations.
  In this paper, we introduce adversarial distributional training (ADT), a novel framework for learning robust models. %Specifically, 
  ADT is formulated as a minimax optimization problem, where the inner maximization aims to learn an adversarial distribution to characterize the potential adversarial examples around a natural one under an entropic regularizer, and the outer minimization aims to train robust models by minimizing the expected loss over the worst-case adversarial distributions. %We augment the inner objective with an entropic regularizer to prevent the adversarial distribution from degeneration and
  Through a theoretical analysis, we develop a general algorithm for solving ADT, and present three approaches for parameterizing the adversarial distributions, ranging from the typical Gaussian distributions to the flexible implicit ones. Empirical results on several benchmarks validate the effectiveness of ADT compared with the state-of-the-art AT methods.
\end{abstract}

%\vspace{-1.5ex}
\section{Introduction}
%\vspace{-0.5ex}
While recent breakthroughs in deep neural networks (DNNs) have led to substantial success in a wide range of fields~\cite{Goodfellow-et-al2016},
%including computer vision \cite{krizhevsky2012imagenet,he2016deep}, speech recognition \cite{Graves2013Speech}, natural language processing \cite{devlin2018bert}, etc.
%However, DNN models are vulnerable to adversarial examples \cite{szegedy2013intriguing,goodfellow2014explaining}, which are indistinguishable from natural examples but make a model produce erroneous predictions. The adversarial vulnerability of DNNs limits their practical applicability for various security-sensitive applications, such as autonomous driving, healthcare, and finance.
%Therefore, it is essential to develop robust DNN models that are resistant to adversarial examples.
DNNs also exhibit adversarial vulnerability to small perturbations around the input~\cite{szegedy2013intriguing,goodfellow2014explaining}.
Due to the security threat, considerable efforts have been devoted to improving the adversarial robustness of DNNs~\cite{goodfellow2014explaining,kurakin2016adversarial,liao2018defense,madry2017towards,wong2018provable,pang2018towards,xie2019feature,pang2019improving,zhang2019theoretically}. 
Among them, adversarial training (AT) is one of the most effective techniques~\cite{Athalye2018Obfuscated,dong2020benchmarking}.
AT can be formulated as a minimax optimization problem~\cite{madry2017towards}, where the inner maximization aims to find an adversarial example that maximizes the classification loss for a natural one, while the outer minimization aims to train a robust classifier using the generated adversarial examples.
To solve the non-concave and typically intractable inner maximization problem approximately, several adversarial attack methods can be adopted, such as fast gradient sign method (FGSM)~\cite{goodfellow2014explaining} and projected gradient descent (PGD) method~\cite{madry2017towards}.

%However, AT also suffers from two generalization problems. First, the generalization gap for robust accuracy between training and test data is significantly larger than the generalization gap for natural accuracy. 
%\citet{schmidt2018adversarially} have shown that adversarially robust generalization requires larger sample complexity than natural generalization.
However, existing AT methods usually solve the inner maximization problem based on a specific attack algorithm, some of which can result in poor generalization for other unseen attacks \emph{under the same threat model}~\cite{song2019improving}\footnote{It should be noted that we consider the generalization problem across attacks \emph{under the same threat model}, rather than studying the generalization ability \emph{across different threat models}~\cite{hendrycks2019benchmarking,engstrom2019exploring,tramer2019adversarial}.}.
For example, defenses trained on the FGSM adversarial examples, without random initialization or early stopping~\cite{wong2020fast}, are vulnerable to multi-step attacks~\cite{kurakin2016adversarial,tramer2017ensemble}.
Afterwards, recent methods~\cite{zhang2019defense,xiao2020enhancing} can achieve the state-of-the-art robustness against the commonly used attacks (e.g., PGD), but they can still be defeated by others~\cite{lin2019feature,tramer2020adaptive}.
It indicates that these defenses probably cause gradient masking~\cite{tramer2017ensemble,Athalye2018Obfuscated,uesato2018adversarial}, and can be fooled by stronger or adaptive attacks. 
%Therefore, there is still a \emph{generalization problem across attacks} in the existing AT methods.

Moreover, a single attack algorithm could be insufficient to explore the space of possible perturbations. PGD addresses this issue by using random initialization, however the adversarial examples crafted by PGD with random restarts probably lie together and lose diversity~\cite{tashiro2020output}. 
As one key to the success of AT is how to solve the inner maximization problem, other methods perform training against multiple adversaries~\cite{tramer2017ensemble,jang2019adversarial}, which can be seen as more exhaustive approximations of the inner problem~\cite{madry2017towards}. Nevertheless, there still lacks a formal characterization of multiple, diverse adversaries.

%However, the performance of the state-of-the-art AT methods is still unsatisfactory by showing only a modest level of adversarial robustness.
%A line of work improves AT by exploiting more data~\cite{hendrycks2019using,alayrac2019labels,carmon2019unlabeled}.
%Nevertheless, in this work we focus on designing a more effective training mechanism to tackle this problem, which is orthogonal to using more data.

%\footnote{This argument is recently challenged by \citet{wong2020fast}. However, their proposed \textit{fast adversarial training} supplements FGSM adversarial training with several tricks, such as random initialization, cyclic learning rate, and early stop. So the basic FGSM adversarial training \cite{kurakin2016adversarial} still has the generalization issue for multi-step attacks.}.\tianyu{Too long footnote. We can simply refer to 'trained on naive FGSM' or 'trained on naive FGSM without random initialization'}
%Although PGD is shown to be a universal ``first-order'' adversary \cite{madry2017towards}, i.e., models robust against PGD will be robust against other first-order gradient-based attacks,
%Although PGD is commonly adopted to evaluate the robustness of AT-based defenses,

%\footnote{A discussion on it can be found at \url{https://openreview.net/forum?id=Syejj0NYvr&noteId=rkeBhuBMjS}.}. %\hangx{we can give a brief summarization about it? why PGD can be universal and why it can still be attacked.}

To mitigate the aforementioned issues and improve the model robustness against a wide range of adversarial attacks, in this paper we present \textbf{adversarial distributional training (ADT)}, a novel framework that explicitly models the adversarial examples around a natural input using a distribution. Subsuming AT as a special case, ADT is formulated as a minimax problem, where the inner maximization aims to find an adversarial distribution for each natural example by maximizing the expected loss over this distribution, while the outer minimization aims to learn a robust classifier by minimizing the expected loss over the worst-case adversarial distributions. 
%ADT naturally subsumes AT as a special case. 
To keep the adversarial distribution from collapsing into a Delta one, we explicitly add an entropic regularization term into the objective, making the distribution capable of characterizing heterogeneous adversarial examples.
%around the natural one.
%Compared with the vanilla AT that generates an adversarial example for each input using a single attack, ADT learns an adversarial distribution that assigns probabilities for a large adversarial region. This region potentially contains adversarial examples given by various attack methods, such that minimizing the expected loss over this region would lead to a better generalization ability across attacks.
%Moreover, the region covered by the adversarial distribution can help to learn a smooth and flattened loss surface in the input space, as shown in Fig.~\ref{fig:loss}.
%the adversarial region could even contain adversarial examples that are unreachable by the first-order adversaries. \hangx{better to give more justifications to convince the reviewers.} 
%Consequently, ADT can improve the overall robustness of the trained models over AT, without sacrificing natural accuracy.

Through a theoretical analysis, %on how to solve the minimax optimization problem in ADT, which indicates 
we show that the minimax problem of ADT can be solved sequentially similar to AT~\cite{madry2017towards}.
We implement ADT by parameterizing the adversarial distributions with trainable parameters, with three concrete examples ranging from the classical Gaussian distributions to the very flexible implicit density models.
%As concrete examples, we present three approaches to specifying the parameterization techniques and learning strategies, respectively. 
Extensive experiments on the CIFAR-10~\cite{krizhevsky2009learning}, CIFAR-100~\cite{krizhevsky2009learning}, and SVHN~\cite{netzer2011reading} datasets validate the effectiveness of our proposed methods on building robust deep learning models, compared with the alternative state-of-the-art AT methods.%\junz{shall we consider including results on ImageNet?}

%The advantages of learning an adversarial distribution rather than generating a single adversarial example is two-fold. 
%First, as different attacks can generates different adversarial examples, we directly model them as an adversarial distribution that could cover most of them. Training with such a distribution can confer robustness to many different attacks, which helps to solve the generalization problem between attacks.
%Second, learning an adversarial distribution can cover some adversarial regions around the natural example, which are not reachable by simple gradient-based attacks (e.g., PGD). Therefore, the model can learning from diverse adversarial examples drawn from a  

%\vspace{-1ex}
\section{Proposed method}
%\vspace{-1ex}
In this section, we first introduce the background of adversarial training (AT), then detail adversarial distributional training (ADT) framework, and finally provide a general algorithm for solving ADT. 

%\vspace{-1ex}
\subsection{Adversarial training}
%\vspace{-1ex}
Adversarial training has been widely studied to improve the adversarial robustness of DNNs.
Given a dataset $\mathcal{D}=\{(\vect{x}_i,y_i)\}_{i=1}^n$ of $n$ training samples with $\vect{x}_i \in \mathbb{R}^d$ and $y_i\in\{1,...,C\}$ being the natural example and the true label, AT can be formulated as a minimax optimization problem~\cite{madry2017towards} as 
\begin{equation}
\label{eq:at}
    \min_{\vect{\theta}}\frac{1}{n}\sum_{i=1}^{n}\max_{\vect{\delta}_i\in\mathcal{S}}\mathcal{L}(f_{\vect{\theta}}(\vect{x}_i + \vect{\delta}_i), y_i),
\end{equation}
where $f_{\vect{\theta}}$ is the DNN model with parameters $\vect{\theta}$ that outputs predicted probabilities over all classes, $\mathcal{L}$ is a loss function (e.g., cross-entropy loss), and $\mathcal{S} = \{{\vect{\delta}}: \|{\vect{\delta}}\|_{\infty}\leq\epsilon\}$ is a perturbation set with $\epsilon>0$.
This is the $\ell_{\infty}$ threat model widely studied before and what we consider in this paper. Our method can also be extended to other threat models (e.g., $\ell_2$ norm), which we leave to future work.

This minimax problem is usually solved sequentially, i.e., adversarial examples are crafted by solving the inner maximization first, and then the model parameters are optimized based on the generated adversarial examples.
Several attack methods can be used to solve the inner maximization problem approximately, such as FGSM~\cite{goodfellow2014explaining} or PGD~\cite{madry2017towards}. For example, PGD takes multiple gradient steps as
%The fast gradient sign method (FGSM) \cite{goodfellow2014explaining} generates the adversarial perturbation $\vect{\delta}_i^*$ using the loss gradient with respect to the input as
%\begin{equation}
%    \vect{\delta}_i^*= \epsilon\cdot\mathrm{sign}(\nabla_{\vect{x}}\mathcal{L}(f_{\vect{\theta}}(\vect{x}_i),y_i)).
%\end{equation}
%The projected gradient descent method (PGD)~\cite{madry2017towards} takes multiple gradient steps as
\begin{equation} 
    \vect{\delta}_i^{t+1}=\Pi_{\mathcal{S}}\big(\vect{\delta}_i^t + \alpha\cdot\mathrm{sign}(\nabla_{\vect{x}}\mathcal{L}(f_{\vect{\theta}}(\vect{x}_i + \vect{\delta}_i^t),y_i))\big),
\end{equation}
where $\vect{\delta}_i^t$ is the adversarial perturbation at the $t$-th step, $\Pi(\cdot)$ is the projection function, and $\alpha$ is a small step size.
$\vect{\delta}_i^0$ is initialized uniformly in $\mathcal{S}$. $\vect{\delta}_i^t$ will converge to a local maximum eventually. 
%For the $T$-step PGD, the final adversarial perturbation is given by $\vect{\delta}_i^*=\vect{\delta}_i^T$.

%\vspace{-1ex}
\subsection{Adversarial distributional training}
%\vspace{-1ex}
\label{sec:adt}

As we discussed, though effective, AT is not problemless. %encounters some problems. 
%The performance of the state-of-the-art AT methods is far from satisfactory. And their generalization ability across attacks is poor. 
AT with a specific attack possibly leads to overfitting on the attack pattern~\cite{kurakin2016adversarial,tramer2017ensemble,zhang2019defense,xiao2020enhancing}, which hinders the trained models from defending against other attacks.
And a single attack algorithm may be unable to explore all possible perturbations in the high-dimensional space, which could result in unsatisfactory robustness performance~\cite{tramer2017ensemble,jang2019adversarial}.
%on the other hand, the manipulated adversarial examples have zero support in the input space, and hence optimizing classification loss on these points in a few iterations can hardly enforce the local smoothness around the natural ones.
%\junz{As in abstract, before giving the math details, we need a good motivation for the moving from AT to distribution, e.g., by analyzing the weakness of AT. Match formulation is a rigorous form of our reasoning. Even without the math, readers should be able to understand our key contributions.}

To alleviate these problems, we propose to capture the distribution of adversarial perturbations around each input instead of only finding a locally most adversarial point for more generalizable adversarial training, called \textbf{adversarial distributional training (ADT)}. 
%\hangx{why it can improve the generalization?}
In particular, we model the adversarial perturbations around each natural example $\vect{x}_i$ by a distribution $p(\vect{\delta}_i)$, whose support is contained in $\mathcal{S}$.
%Instead of training on a single adversarial example as in Eqn.~\eqref{eq:at}, we leverage the adversarial distribution for training in ADT.
Based on this, ADT is formulated as a distribution-based minimax optimization problem as 
\begin{equation}
\label{eq:adt}
    \min_{\vect{\theta}}\frac{1}{n}\sum_{i=1}^{n}\max_{p(\vect{\delta}_i)\in\mathcal{P}}\mathbb{E}_{p(\vect{\delta}_i)}\big[ \mathcal{L}(f_{\vect{\theta}}(\vect{x}_i + \vect{\delta}_i), y_i)\big],
\end{equation}
where $\mathcal{P}=\{p:\mathrm{supp}(p)\subseteq \mathcal{S}\}$ is a set of distributions with support contained in $\mathcal{S}$.
As can be seen in Eq.~\eqref{eq:adt}, the inner maximization aims to learn an adversarial distribution, such that a point drawn from it is likely an adversarial example. And the objective of the outer minimization is to adversarially train the model parameters by minimizing the expected loss over the worst-case adversarial distributions induced by the inner problem. It is noteworthy that AT is a special case of ADT, by specifying the distribution family $\mathcal{P}$ to contain Delta distributions only.

\textbf{Regularizing adversarial distributions.} For the inner maximization of ADT, we can easily see that 
\begin{equation}
     \max_{p(\vect{\delta}_i)\in\mathcal{P}}\mathbb{E}_{p(\vect{\delta}_i)}\big[ \mathcal{L}(f_{\vect{\theta}}(\vect{x}_i + \vect{\delta}_i), y_i)\big] \leq  \max_{\vect{\delta}_i\in\mathcal{S}}\mathcal{L}(f_{\vect{\theta}}(\vect{x}_i + \vect{\delta}_i), y_i).
\end{equation}
It indicates that the optimal distribution by solving the inner problem of ADT will degenerate into a Dirac one.
Hence the adversarial distribution cannot cover a diverse set of adversarial examples, and ADT becomes AT.
To solve this issue, we add an entropic regularization term into the objective as
\begin{equation}
\label{eq:adt-ent}
    \min_{\vect{\theta}}\frac{1}{n}\sum_{i=1}^{n}\max_{p(\vect{\delta}_i)\in\mathcal{P}} \mathcal{J}\big(p(\vect{\delta}_i), \vect{\theta}\big), \; \text{with }
    \mathcal{J}\big(p(\vect{\delta}_i), \vect{\theta}\big) = \mathbb{E}_{p(\vect{\delta}_i)}\big[ \mathcal{L}(f_{\vect{\theta}}(\vect{x}_i + \vect{\delta}_i), y_i)\big] + \lambda\mathcal{H}(p(\vect{\delta}_i)),
\end{equation}
where $\mathcal{H}(p(\vect{\delta}_i))=-\mathbb{E}_{p(\vect{\delta}_i)}[\log p(\vect{\delta}_i)]$ 
is the entropy of $p(\vect{\delta}_i)$, $\lambda$ is a balancing hyperparameter, and $\mathcal{J}\big(p(\vect{\delta}_i), \vect{\theta}\big)$ denotes the overall loss function for notation simplicity.
Note that the entropy maximization is a common technique to increase the support of a distribution in generative modeling~\cite{dai2017calibrating,NIPS2017_7229} or reinforcement learning~\cite{haarnoja2018soft}.
We next discuss why ADT is superior to AT.
%the training are performed in a nested manner: in the inner loop, we update the distributions to capture the adversarial regions in the input space given current model; in the outer loop, the model is optimized to be adversarially robust under the entire perturbation distributions.

\begin{figure}[t]
\centering
\begin{minipage}{0.46\linewidth}
\centering
\includegraphics[width=1.0\linewidth]{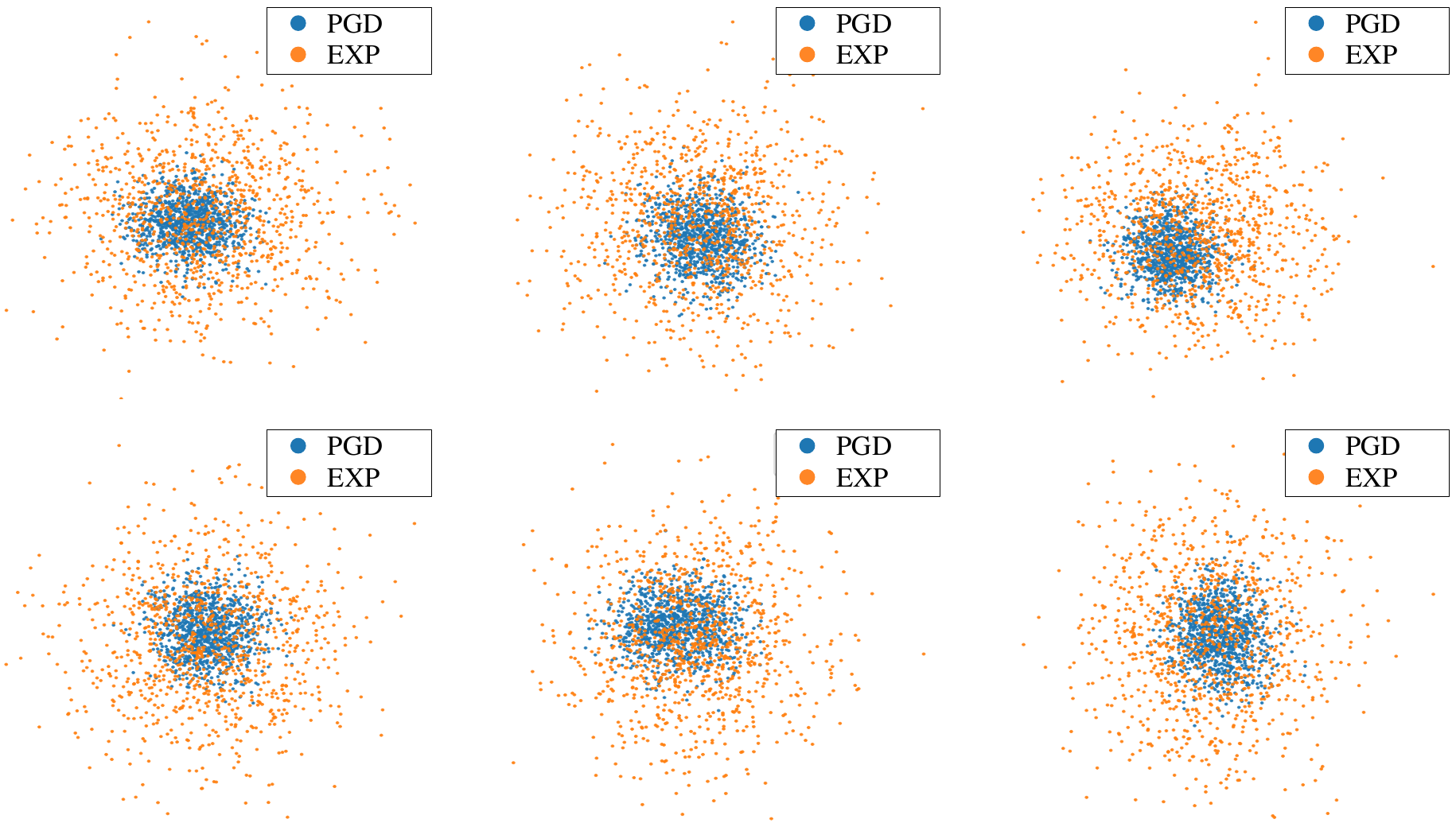}
\caption{Visualization of the adversarial examples generated by PGD with random restarts (blue) and those sampled from the adversarial distribution learned by ADT\textsubscript{EXP} (orange). Each subfigure corresponds to one randomly selected data point.}
\label{fig:pgd}
\end{minipage}
\hspace{1.ex}
\begin{minipage}{0.515\linewidth}
\centering
\includegraphics[width=1.0\linewidth]{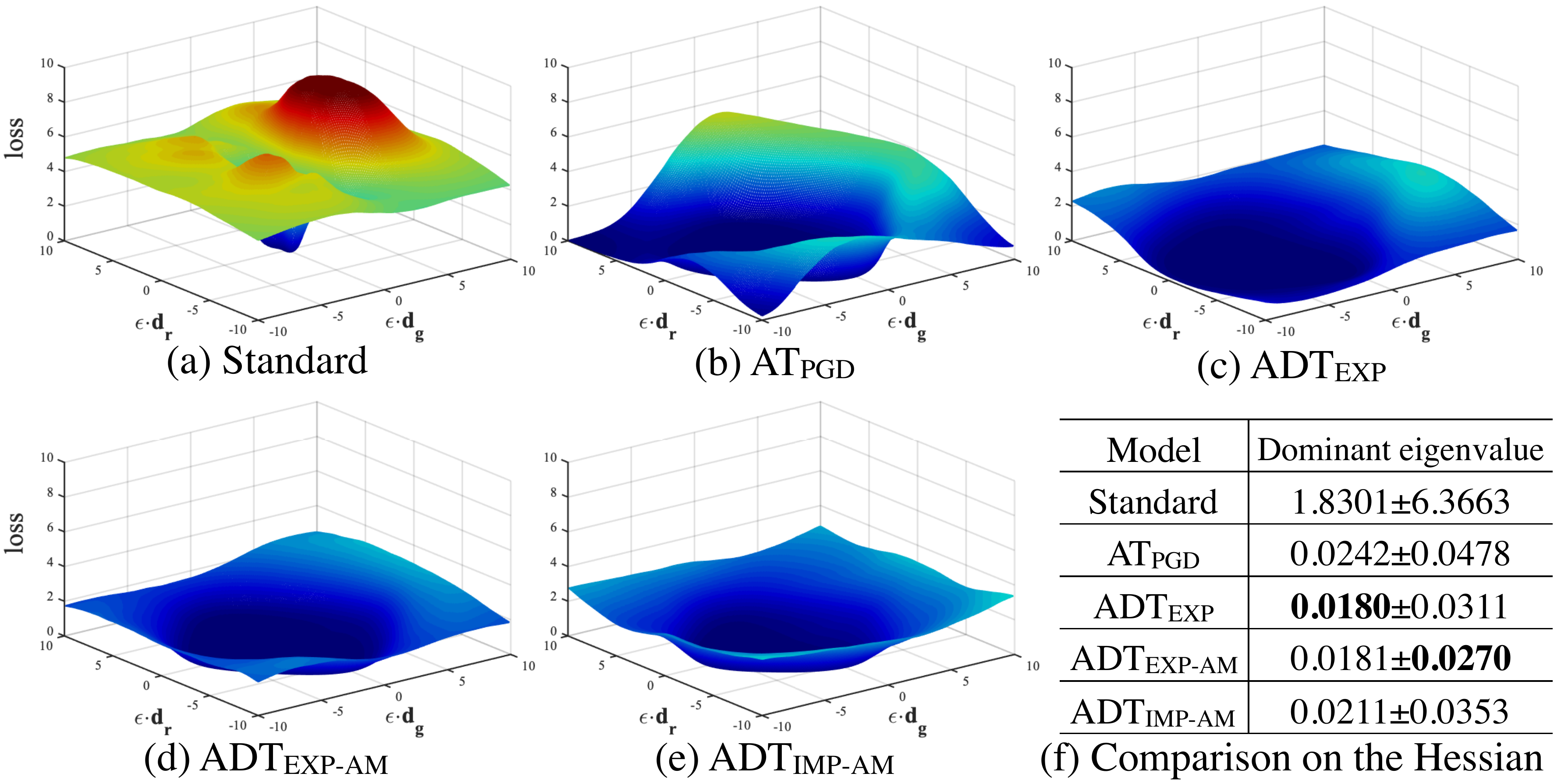}
\caption{Visualization of loss surfaces in the vicinity of an input along the gradient direction
($\vect{d_g}$) and a random direction ($\vect{d_r}$) for various models in (a)-(e). (f) reports the dominant eigenvalue of the Hessian matrix of the classification loss w.r.t. the input. Full details are in Sec.~\ref{sec:ablation}.}
\label{fig:loss}
\end{minipage}
%\vspace{-2ex}
\end{figure}

\subsubsection{Discussion on the superiority of ADT}

The major difference between AT and ADT is that for each natural input $\vect{x}_i$, AT finds a worst-case adversarial example, while ADT learns a worst-case adversarial distribution 
comprising a variety of adversarial examples.
%that assigns probabilities for a relatively large adversarial region contained in $\mathcal{S}$
%\junz{not sure on what this means: when you talk about a distribution over $S$, don't you by default mean a full support over $S$? why a subset of $S$? Is there any special treatment on $S$, compared to AT?}.
%\junz{what do you mean by "large region"? the support of the distribution is the same as in AT? If so, the key difference is that AT finds a single point, while ADT finds a distribution with AT as a special case.}
Because adversarial examples can be generated by various attacks, we expect that those adversarial examples probably lie in the region where the adversarial distribution assigns high probabilities, such that minimizing the expected loss over this distribution can naturally lead to a better generalization ability of the trained classifier across attacks under the same threat model.

Furthermore, as we add an entropic regularizer into the objective~\eqref{eq:adt-ent}, the adversarial distribution is able to better explore the space of possible perturbations and characterize more diverse adversarial examples compared with a single attack method (e.g., PGD).
To show this, for each data we generate a set of adversarial examples by PGD with random restarts and sample another set of adversarial examples from the adversarial distribution learned by ADT\textsubscript{EXP} (a variant of ADT detailed in Sec.~\ref{sec:adt_gau}), targeted at a standard trained model. Then we can visualize these adversarial examples by projecting them onto the 2D space spanned by the first two eigenvectors given by PCA~\cite{jolliffe1986principal}.
The visualization results of some randomly selected data points in Fig.~\ref{fig:pgd} show that adversarial examples sampled from the adversarial distribution are scattered while those crafted by PGD concentrate together. We further evaluate the diversity of adversarial examples by quantitatively measuring their average pairwise distances. 
The average $\ell_2$ distance of adversarial examples sampled from the adversarial distribution over $100$ test images is $1.95$, which is $1.56$ for PGD. Although the adversarial distributions can characterize more diverse adversarial examples, they have a similar attack power compared with PGD, as later shown in Table~\ref{table:g_attack}.
%Many of the adversarial examples sampled from the adversarial distribution are not reachable by PGD since they are not local maxima of the loss function.
Minimizing the loss on such diverse adversarial examples can consequently help to learn a smoother and more flattened loss surface around the natural examples in the input space, as shown in Fig.~\ref{fig:loss}. Therefore, ADT can improve the overall robustness compared with AT.

\begin{algorithm}[tb]
\small
   \caption{The general algorithm for ADT}
   \label{alg:training}
\begin{algorithmic}[1]
   \Require Training data $\mathcal{D}$, objective function $\mathcal{J}\big(p(\vect{\delta}_i), \vect{\theta}\big)$, the set of perturbation distributions $\mathcal{P}$, training epochs $N$, and learning rate $\eta$.
   \State  Initialize $\vect{\theta}$;
   \For {$\text{epoch}=1$ {\bfseries to} $N$}
   \For {each minibatch $\mathcal{B}\subset\mathcal{D}$}
   \State Obtain $p^*(\vect{\delta}_i)$ for each input $(\vect{x}_i, y_i) \in \mathcal{B}$ by solving $p^*(\vect{\delta}_i) = \argmax_{p(\vect{\delta}_i)\in\mathcal{P}}\mathcal{J}\big(p(\vect{\delta}_i), \vect{\theta}\big)$;
   \State Update $\vect{\theta}$ with stochastic gradient descent $\vect{\theta}\leftarrow\vect{\theta}-\eta\cdot\mathbb{E}_{(\vect{x}_i, y_i)\in\mathcal{B}}\big[\nabla_{\vect{\theta}}\mathcal{J}\big(p^*(\vect{\delta}_i), \vect{\theta}\big)\big]$;
   \EndFor
   \EndFor
\end{algorithmic}
\end{algorithm}

\subsection{A general algorithm for ADT}
%\vspace{-1ex}
To solve minimax problems, Danskin's theorem~\cite{danskin2012theory} states how the maximizers of the inner problem can be used to define the gradients for the outer problem, which is also the theoretical foundation of AT~\cite{madry2017towards}.
However, it is problematic to directly apply Danskin's theorem for solving ADT since the search space $\mathcal{P}$ may not be compact, which is one assumption of this theorem.
As it is non-trivial to perform a theoretical analysis on how to solve ADT, we first lay out the following assumptions.

\begin{assumption}\label{ass:1}
The loss function $\mathcal{J}\big(p(\vect{\delta}_i), \vect{\theta}\big)$ is continuously differentiable w.r.t. $\vect{\theta}$.
\end{assumption}
Assumption~\ref{ass:1} is also made in~\cite{madry2017towards} for AT. Although the loss function is not completely continuously differentiable due to the ReLU layers, the set of discontinuous points has measure zero, such that it is assumed not to be an issue in practice~\cite{madry2017towards}.

\begin{assumption}\label{ass:2}
Probability density functions of distributions in $\mathcal{P}$ are bounded and equicontinuous.
\end{assumption}

Assumption~\ref{ass:2} puts a restriction on the set of distributions $\mathcal{P}$. We show that the explicit adversarial distributions proposed in Sec.~\ref{sec:adt_gau} satisfy this assumption (in Appendix~\ref{sec:b-1}).
%since arbitrary distributions do not have bounded and equicontinuous density functions. \junz{do the distributions defined in Sec 3 satisfy?} 

%Given these assumptions, we have the following theorem.
%In practice, we parameterize the distributions, which can satisfy this assumption.

\begin{theorem}\label{the:1}
Suppose Assumptions~\ref{ass:1} and \ref{ass:2} hold. We define $\rho(\vect{\theta})=\max_{p(\vect{\delta}_i)\in\mathcal{P}}\mathcal{J}\big(p(\vect{\delta}_i), \vect{\theta}\big)$, and $\mathcal{P}^*(\vect{\theta}) = \{p(\vect{\delta}_i)\in\mathcal{P}:\mathcal{J}\big(p(\vect{\delta}_i), \vect{\theta}\big)=\rho(\vect{\theta})\}$. Then $\rho(\vect{\theta})$ is directionally differentiable, and its directional derivative along the direction $\vect{v}$ satisfies
\begin{equation}
    \rho'(\vect{\theta};\vect{v}) = \sup_{p(\vect{\delta}_i) \in \mathcal{P}^*(\vect{\theta})}\vect{v}^\top\nabla_{\vect{\theta}}\mathcal{J}\big(p(\vect{\delta}_i), \vect{\theta}\big).
\end{equation}
Particularly, when $\mathcal{P}^*(\vect{\theta}) = \{p^*(\vect{\delta}_i)\}$ only contains one maximizer, $\rho(\vect{\theta})$ is differentiable at $\vect{\theta}$ and
\begin{equation}
    \nabla_{\vect{\theta}}\rho(\vect{\theta}) = \nabla_{\vect{\theta}}\mathcal{J}\big(p^*(\vect{\delta}_i), \vect{\theta}\big).
\end{equation}
\end{theorem}
The complete proof of Theorem~\ref{the:1} is deferred to Appendix~\ref{sec:b-1}. Theorem~\ref{the:1} provides us a general principle for training ADT, by first solving the inner problem and then updating the model parameters along the gradient direction of the loss function at the global maximizer of the inner problem, in a sequential manner similar to AT~\cite{madry2017towards}.
We provide the general algorithm for ADT in Alg.~\ref{alg:training}.
Analogous to AT, the global maximizer of the inner problem cannot be solved analytically. Therefore, we propose three different approaches to obtain approximate solutions, as introduced in Sec.~\ref{sec:method}. 
Although we cannot reach the global maximizer of the inner problem,
%the exact assumptions in our theorem do not hold,\junz{seems a gap between theory and practice, is there some special case to meet the assumption that still work reasonably well?} 
our experiments suggest that we can reliably solve the minimax problem~\eqref{eq:adt-ent} by our algorithm.

%\vspace{-1ex}
\section{Parameterizing adversarial distributions}
\begin{figure}
\centering
\begin{minipage}{0.435\textwidth}
\centering
%\vspace{-2ex}
\includegraphics[width=1.0\textwidth]{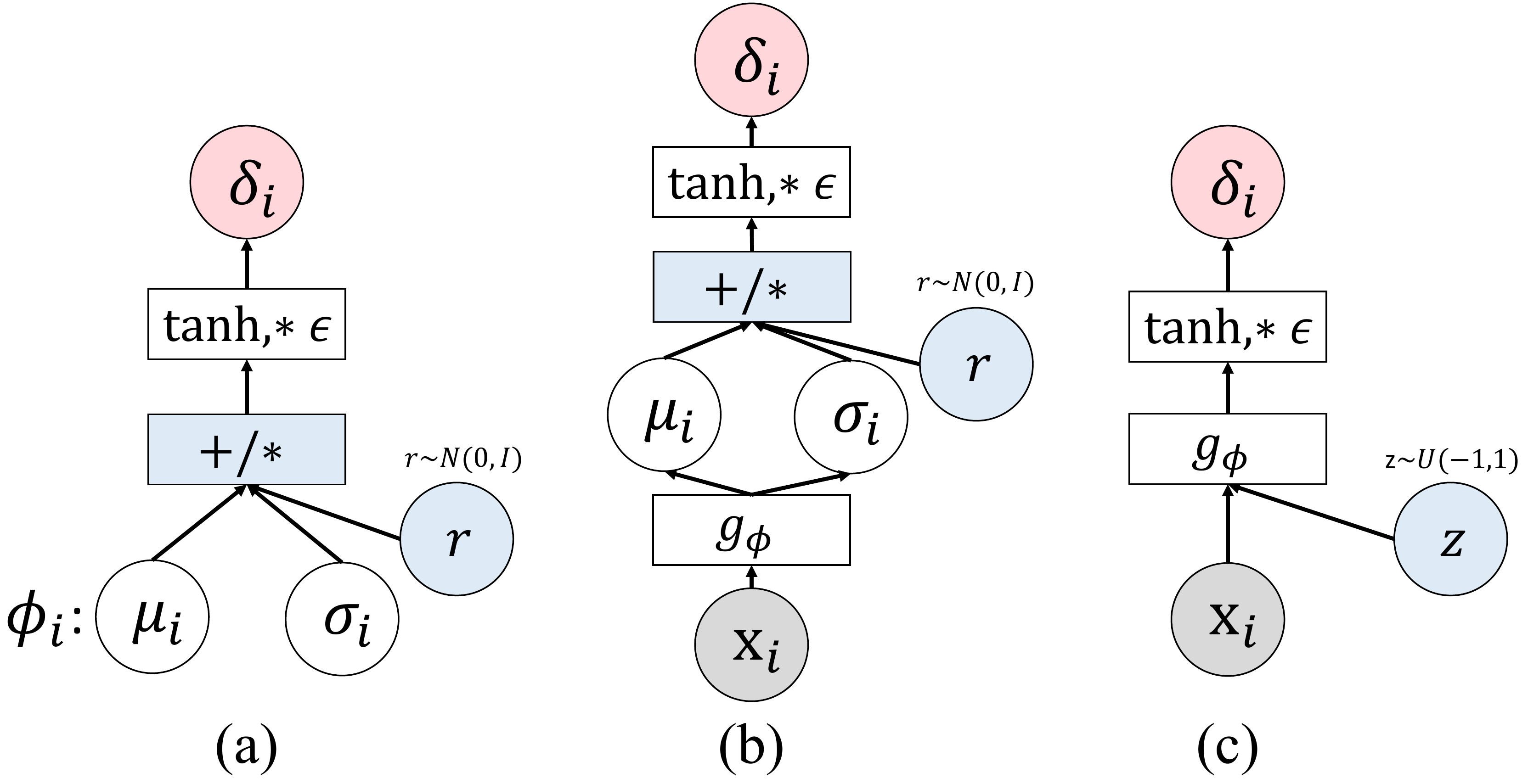}
\end{minipage}
\hspace{2ex}
\begin{minipage}{0.53\textwidth}
\captionsetup{font={small}}
\caption{An illustration of the three different approaches to parameterize the distributions of adversarial perturbations. (a) ADT\textsubscript{EXP}: the explicit adversarial distribution $p_{\vect{\phi}_i}(\vect{\delta}_i)$ is defined by transforming $\mathcal{N}(\vect{\mu}_i, \mathrm{diag}(\vect{\sigma}_i^2))$ via $\tanh$ followed by a multiplication with $\epsilon$. (b) ADT\textsubscript{EXP-AM}: we amortize the explicit adversarial distributions by a neural network $g_{\vect{\phi}}$ taking $\vect{x}_i$ as input. (c) ADT\textsubscript{IMP-AM}: we define the implicit adversarial distributions by inputting an additional random variable $\vect{z}\sim \mathrm{U}(-1,1)$ to the network $g_{\vect{\phi}}$.}
%\hangx{more explanations for each algorithm, and ($tanh$, $\ast\epsilon$) is a little bit confusing}}
\label{fig:approaches}
\end{minipage}
\vspace{0ex}
\end{figure}

%\vspace{-1ex}
\label{sec:method}
At the core of ADT lie the solutions of the inner maximization problem of Eq.~\eqref{eq:adt-ent}. The basic idea is to parameterize the adversarial distributions with trainable parameters $\vect{\phi}_i$. With the parameterized $p_{\vect{\phi}_i}(\vect{\delta}_i)$, the inner problem is converted into maximizing the expected loss w.r.t. $\vect{\phi}_i$. % we directly optimize $\phi$ so that the samples from the distributions have relatively high classification error averagely. 
In the following, we present the parametrizations and learning strategies of three different approaches, respectively.
We provide an overview of these approaches in Fig.~\ref{fig:approaches}.
%\junz{it seems the three parametrizations produce mixing results in experiments, be careful on the statement, e.g., provide suggestions on which one to choose?}\junz{if not really needed, may choose to shrink or move one of them to appendix, and give more space to eleborate the key change of moving from AT to ADT (e.g., diversity of adversarial samples)... keep the main message clear and avoid a too-heavy load for readers.}

%\vspace{-1ex}
\subsection{ADT\textsubscript{EXP}: explicit modeling of adversarial perturbations}
%\vspace{-1ex}
\label{sec:adt_gau}

A natural way to model adversarial perturbations around an input data is using a distribution with an explicit density function. We name ADT with EXPlicit adversarial distributions as ADT\textsubscript{EXP}.
%Typically, we assume $\vect{\delta}_i$ follows a multivariate Gaussian with a diagonal covariance. But recall that its support should be contained in the compact set $\mathcal{S} = \{{\vect{\delta}}: \|{\vect{\delta}}\|_{\infty}\leq\epsilon\}$. Therefore, we take the transformation of random variable approach to define $p_{\vect{\phi}_i}(\vect{\delta}_i)$ as
To define a proper distribution $p_{\vect{\phi}_i}(\vect{\delta}_i)$ on $\mathcal{S}$, we take the transformation of random variable approach as
\begin{equation}
\label{eq:explicit}
    \vect{\delta}_i = \epsilon\cdot\tanh (\vect{u}_i), \quad \vect{u}_i \sim \mathcal{N}(\vect{\mu}_i, \mathrm{diag}(\vect{\sigma}_i^2)),
\end{equation}
where $\vect{u}_i$ is sampled from a diagonal Gaussian distribution with $\vect{\mu}_i, \vect{\sigma}_i\in\mathbb{R}^d$ as the mean and standard deviation. $\vect{u}_i$ is transformed by a $\tanh$ function and then multiplied by $\epsilon$ to get $\vect{\delta}_i$. We let $\vect{\phi}_i=(\vect{\mu}_i, \vect{\sigma}_i)$ denote the parameters to be learned.  
We sample $\vect{u}_i$ from a diagonal Gaussian mainly for the sake of computational simplicity. But our method is fully compatible with more expressive distributions, such as matrix-variate Gaussians~\cite{louizos2016structured} or multiplicative normalizing flows~\cite{louizos2017multiplicative}, and we leave using them to future work.
Given Eq.~\eqref{eq:explicit}, the inner problem of Eq.~\eqref{eq:adt-ent} becomes
\begin{equation}
%\label{eq:adt_gau}
    \max_{\vect{\phi}_i}\Big\{\mathbb{E}_{p_{\vect{\phi}_i}(\vect{\delta}_i)}\big[ \mathcal{L}(f_{\vect{\theta}}(\vect{x}_i + \vect{\delta}_i), y_i)\big] + \lambda \mathcal{H}(p_{\vect{\phi}_i}(\vect{\delta}_i))\Big\}.
\end{equation}
To solve this, we need to estimate the gradient of the expected loss w.r.t. the parameters $\vect{\phi}_i$. A commonly used method is the low-variance reparameterization trick~\cite{kingma2013auto,blundell2015weight}, which replaces the sampling process of the random variable of interest with the corresponding differentiable transformation.
%which replaces the sampling procedure of $\vect{\delta}_i$ with the
%corresponding differentiable deterministic transformation using a sample of parameter-free noise. 
With this technique, the gradient can be back-propagated from the samples to the distribution parameters directly. In our case, we reparameterize $\vect{\delta}_i$ by $\vect{\delta}_i=\epsilon\cdot\tanh(\vect{u}_i)=\epsilon\cdot\tanh(\vect{\mu}_i + \vect{\sigma}_i\vect{r})$, where $\vect{r}$ is an auxiliary noise variable following the standard Gaussian distribution $\mathcal{N}(\vect{0}, \vect{I})$. Therefore, we can estimate the gradient of $\vect{\phi}_i$ via
%the following gradient estimator of $\vect{\phi}_i$ \yinpeng{Is it called the gradient estimator? Since no estimation is used.}
\begin{equation}
\label{eq:adt_gau_der}
\small
%\begin{split}
    \mathbb{E}_{\vect{r}\sim\mathcal{N}(\vect{0},\vect{I})} \nabla_{\vect{\phi}_i}\Big[\mathcal{L}\big(f_{\vect{\theta}}\big(\vect{x}_i + \epsilon\cdot \tanh(\vect{\mu}_i + \vect{\sigma}_i\vect{r})\big), y_i\big) -\lambda\log p_{\vect{\phi}_i}\big(\epsilon \cdot\tanh(\vect{\mu}_i + \vect{\sigma}_i\vect{r})\big)\Big].
%\end{split}
\end{equation}
%where the first term is the classification loss over the distribution of perturbations, and the second term is $\mathcal{H}(p_{\vect{\phi}_i}(\vect{\delta}_i))$. 
The first term inside is the classification loss with the sampled noise, and the second is the negative log density (i.e., estimation of entropy). It can be calculated analytically (proof in Appendix~\ref{sec:b-2}) as
\begin{equation}\label{eq:11}
\small
    %\log p_{\vect{\phi}_i}(\vect{\delta}_i) = 
    \sum_{j=1}^d\Big(\frac{1}{2}(\vect{r}^{(j)})^2+\frac{\log2\pi}{2}+\log\vect{\sigma}_i^{(j)}+\log\big(1-\tanh(\vect{\mu}_i^{(j)} + \vect{\sigma}_i^{(j)}\vect{r}^{(j)})^2\big)+\log\epsilon\Big),
\end{equation}
where the superscript $j$ denotes the $j$-th element of a vector.

In practice, we approximate the expectation in Eq.~\eqref{eq:adt_gau_der} with $k$ Monte Carlo (MC) samples, and perform $T$ steps of gradient ascent on $\vect{\phi}_i$ to solve the inner problem. After obtaining the optimal parameters $\vect{\phi}_i^*$, we use the adversarial distribution $p_{\vect{\phi}_i^*}(\vect{\delta}_i)$ to update model parameters $\vect{\theta}$.
%that, to learn a robust model, we update the model parameters $\vect{\theta}$ given the learned $p_{\vect{\phi}_i}(\vect{\delta}_i)$ which %coherently \yinpeng{why coherently?} 
%characterizes the adversarial perturbations around the natural example $\vect{x}_i$. 

%Intuitively, the optimization process of the perturbation distributions makes an analogy with multi-step adversarial attack methods (e.g., PGD): the perturbation (distribution) of one example is updated to be adversarial asymptotically. However, it has been shown that PGD-like approaches which adopts first-order gradient ascent are likely to be stuck in extensive saddle points cite{jin}. In contrast, by importing the randomness into the produced perturbation samples in a principled way, ADT has a relatively high potential to escape from these saddle points. This may help to find more adversarial perturbations and be beneficial to the training of the model moderately (do we need empirical proof). \yinpeng{This paragraph can be removed since the advantage is illustrated in Section 2.2.}

%\vspace{-1ex}
\subsection{ADT\textsubscript{EXP-AM}: amortizing the explicit adversarial distributions}
%\vspace{-1ex}
Although the aforementioned method in Sec.~\ref{sec:adt_gau} provides a simple way to learn explicit adversarial distributions for ADT, it needs to learn the distribution parameters for each input and then brings prohibitive computational cost. Compared with PGD-based AT which constructs adversarial examples by $T$ steps PGD~\cite{madry2017towards}, ADT\textsubscript{EXP} is approximately $k$ times slower since the gradient of $\vect{\phi}_i$ is estimated by $k$ MC samples in each step.
In this subsection, we propose to amortize the inner optimization of ADT\textsubscript{EXP}, to develop a more feasible and scalable training method.
We name ADT with the AMortized version of EXPlicit adversarial distributions as ADT\textsubscript{EXP-AM}. %\zhijie{Is the term AMortization right?}

%In this subsection, we amortize the inner optimization of raw ADT, to develop a feasible and scalable adversarial distributional training way, which we refer to as ADT-a. The motivation is immediately obvious: within the inner loop of raw ADT, we have to solve $n$ optimization problems, which may bring prohibitive computational cost, especially when the number of training data $n$ is large. What's more, the $n$ inner problems are inherently related by the same attack objective and target model, so it is a natural choice to convert them into a single learning problem via amortized inference (may not correct and need cite).

Instead of learning the distribution parameters for each data $\vect{x}_i$, we opt to learn a mapping $g_{\vect{\phi}}:\mathbb{R}^d \rightarrow \mathcal{P}$, which defines the adversarial distribution for each input in a conditional manner $p_{\vect{\phi}}(\vect{\delta}_i|\vect{x}_i)$. 
%An appropriate and effective instantiation of this distribution is 
We instantiate $g_{\vect{\phi}}$ by a conditional generator network. It takes a natural example $\vect{x}_i$ as input, and outputs the parameters $(\vect{\mu}_i, \vect{\sigma}_i)$ of its corresponding explicit adversarial distribution, which is also defined by Eq.~\eqref{eq:explicit}.
The advantage of this method is that the generator network can potentially learn common structures of the adversarial perturbations, which can generalize to other training samples~\cite{baluja2017adversarial,poursaeed2018generative}. It means that we do not need to optimize $\vect{\phi}$ excessively on each data $\vect{x}_i$, which can accelerate training.

%\vspace{-1ex}
\subsection{ADT\textsubscript{IMP-AM}: implicit modeling of adversarial perturbations}\label{sec:3-3}
%\vspace{-1ex}
Since the underlying distributions of adversarial perturbations have not been figured out yet and could be different across samples, it is hard to specify a proper explicit distribution of adversarial examples, which may lead to the underfitting problem. 
To bypass this, we resort to implicit distributions (i.e., distributions without tractable probability density functions but can still be sampled from), which have shown promising results recently~\cite{goodfellow2014generative,shi2018kernel,shi2018spectral}, 
particularly in modeling complex high-dimensional data~\cite{radford2015unsupervised,isola2017image}. The major advantage of implicit distributions is that they are not confined to provide explicit densities, which improves the flexibility inside the sampling process.% and {brings benefits to capture the complex correlations in the high-dimensional space.}%\hangx{is it necessary to mention it?}

Based on this, we propose to use the implicit distributions to characterize the adversarial perturbations. Considering the priority of amortized optimization, we learn a generator $g_{\vect{\phi}}: \mathbb{R}^{d_z} \times \mathbb{R}^d \rightarrow \mathbb{R}^d$ which implicitly defines a conditional distribution $p_{\vect{\phi}}(\vect{\delta}_i|\vect{x}_i)$ by $\vect{\delta}_i=g_{\vect{\phi}}(\vect{z};\vect{x}_i)$, where $\vect{x}_i$ is a natural input and $\vect{z}\in \mathbb{R}^{d_z}$ is a random noise vector. Typically, $\vect{z}$ is sampled from a prior $p(\vect{z})$ such as the standard Gaussian or uniform distributions as in the generative adversarial networks (GANs)~\cite{goodfellow2014generative}. 
In this work, we sample $\vect{z}$ from a uniform distribution $\mathrm{U}(-1,1)$. We refer to this approach as ADT\textsubscript{IMP-AM}.
A practical problem remains unaddressed is that the entropy of the implicit distributions cannot be estimated exactly as we have no access to the density $p_{\vect{\phi}}(\vect{\delta}_i|\vect{x}_i)$. We instead maximize the variational lower bound of the entropy~\cite{dai2017calibrating} for its simplicity and success in GANs~\cite{NIPS2017_7229}.
We provide full technical details of ADT\textsubscript{EXP-AM} and ADT\textsubscript{IMP-AM}, and training algorithms of the three methods in Appendix~\ref{sec:a}.

%\vspace{-1.2ex}
\section{Related work}
%\vspace{-1.2ex}
Adversarial machine learning is an emerging research topic with various attack and defense methods being proposed~\cite{goodfellow2014explaining,Kurakin2016,carlini2017towards,Dong2017,liao2018defense,madry2017towards,wong2018provable,cheng2019improving,pang2019improving}.
Besides PGD-based AT~\cite{madry2017towards}, recent improvements upon it include designing new losses~\cite{zhang2019theoretically,mao2019metric,pang2019rethinking,qin2019adversarial} or network architecture~\cite{xie2019feature}, accelerating the training procedure~\cite{shafahi2019adversarial,zhang2019you,wong2020fast}, and exploiting more data~\cite{hendrycks2019using,alayrac2019labels,carmon2019unlabeled,zhai2019adversarially}. %We propose a new training paradigm, which is orthogonal to those prior works.
%The threat of adversarial examples has motivated extensive research on building robust models to defend against adversarial attacks, including defensive distillation \cite{PapernotDistillation2016}, input denoising \cite{guo2017countering,liao2018defense}, randomization \cite{xie2017mitigating}, adversarial training \cite{goodfellow2014explaining,kurakin2016adversarial,madry2017towards}, and certified defenses \cite{sinha2018certifying,wong2018provable,wong2018scaling}. Among these defense methods, adversarial training (AT) demonstrates superior empirical robustness without causing obfuscated gradients \cite{Athalye2018Obfuscated}. Recent work further improves AT for better robustness \cite{zhang2019theoretically,wang2019convergence,alayrac2019labels,carmon2019unlabeled}. Our work is devoted to mitigating the generalization problems of AT via adversarial distributional training (ADT), which is also related to some previous work in the literature.
%Our work is related to some previous work in the literature.

Learning the distributions of adversarial examples has been studied before, mainly for black-box adversarial attacks. An adversarial example can be searched over a distribution~\cite{ilyas2018black,li2019nattack}, similar to the inner problem of Eq.~\eqref{eq:adt}. But their gradient estimator based on natural evolution strategy exhibits very high variance~\cite{kingma2013auto} compared with ours in Eq.~\eqref{eq:adt_gau_der}, since our methods are based on the white-box setting (i.e., compute the gradient) rather than the black-box setting. To the best of our knowledge, we are the first to train robust models by learning the adversarial distributions.

In this work, we adopt a generator network to amortize the adversarial distributions for accelerating the training process. 
There also exists previous work on adopting generator-based approaches for adversarial attacks and defenses~\cite{baluja2017adversarial,poursaeed2018generative,xiao2018generating}. The inner maximization problem of AT can be solved by generating adversarial examples using a generator network~\cite{wang2019direct,chen2018learning}, which is similar to our work. 
The essential difference is that they still focus on the minimax formulation~\eqref{eq:at} of AT, while we propose a novel ADT framework in Eq.~\eqref{eq:adt}. We empirically compare our method with~\cite{chen2018learning} in Appendix~\ref{sec:d-5}.

Adversarial robustness is also related to robustness to certain types of input-agnostic distributions~\cite{ford2019adversarial}. A classifier robust to Gaussian noise can be turned into a new smoothed classifier that is certifiably robust to $\ell_2$ adversarial examples~\cite{cohen2019certified}. \citet{salman2019provably} further employ adversarial training to improve the certified robustness of randomized smoothing, whereas our method belongs to empirical defenses, aiming to train a robust classifier with the input-dependent adversarial distributions.

The proposed ADT framework is essentially different from a seemingly similar concept --- distributionally robust optimization (DRO) \cite{ben2013robust,esfahani2018data,sinha2018certifying}.
DRO seeks a model that is robust against changes in data-generating distribution, by training on the worst-case data distribution under a probability measure. DRO is related to AT with the Wasserstein distance \cite{sinha2018certifying,staib2017distributionally}. However, ADT does not model the changes in data distribution but aims to find an adversarial distribution for each input. 
%\hangx{move to the related work}

%\vspace{-1.2ex}
\section{Experiments}
%\vspace{-1.2ex}
%To empirically validate the effectiveness of ADT on improving the adversarial robustness of DNNs, we perform extensive experiments on several benchmark datasets.

\textbf{Experimental settings and implementation details.}\footnote{Code is available at \url{https://github.com/dongyp13/Adversarial-Distributional-Training}.} We briefly introduce the experimental settings here, and leave the full details in Appendix~\ref{sec:c}. \textbf{(A) Datasets:} We perform experiments on the CIFAR-10~\cite{krizhevsky2009learning}, CIFAR-100~\cite{krizhevsky2009learning}, and SVHN~\cite{netzer2011reading} datasets. The input images are normalized to $[0,1]$. We set the perturbation budget $\epsilon=8/255$ on CIFAR, and $\epsilon=4/255$ on SVHN as in~\cite{carmon2019unlabeled}. \textbf{(B) Network Architectures:} We use a Wide ResNet (WRN-28-10) model~\cite{zagoruyko2016wide} as the classifier in most of our experiments following~\cite{madry2017towards}.
For the generator network used in ADT\textsubscript{EXP-AM} and ADT\textsubscript{IMP-AM}, we adopt a popular image-to-image architecture with residual blocks~\cite{johnson2016perceptual,zhu2017unpaired}.
%We also employ a 5-layer CNN to instantiate $q_{\vect{\psi}}$ in ADT\textsubscript{IMP-AM}. 
\textbf{(C) Training Details:}
We adopt the cross-entropy loss as $\mathcal{L}$ in our objective~\eqref{eq:adt-ent}. 
We set $\lambda=0.01$ for the entropy term, and leave the study of the effects of $\lambda$ in Sec.~\ref{sec:ablation}.
For ADT\textsubscript{EXP}, we adopt Adam \cite{Kingma2014} for optimizing $\vect{\phi}_i$ with the learning rate $0.3$, the optimization steps $T=7$, and the number of MC samples in each step $k=5$.
For ADT\textsubscript{EXP-AM} and  ADT\textsubscript{IMP-AM}, we use $k=1$ for each data for gradient estimation.
\textbf{(D) Baselines:} We adopt two primary baselines: 1) standard training on the natural images (\textbf{\emph{Standard}}); 2) AT on the PGD adversarial examples (\textbf{\emph{AT\textsubscript{PGD}}})~\cite{madry2017towards}. %These baselines are trained with the same parameters specified above. For training AT\textsubscript{PGD}, we perform PGD with $T=7$ steps, and step size $\alpha=\nicefrac{\epsilon}{4}$, which are the same as in \citet{madry2017towards}.
On CIFAR-10, we further incorporate: 1) the pretrained AT\textsubscript{PGD} model (\textbf{\emph{AT\textsubscript{PGD}$^{\dagger}$}}) released by~\cite{madry2017towards}; 2) AT on the targeted FGSM adversarial examples (\textbf{\emph{AT\textsubscript{FGSM}}})~\cite{kurakin2016adversarial}; 3) adversarial logit pairing (\textbf{\emph{ALP}})~\cite{kannan2018adversarial}; and 4) feature scattering-based AT (\textbf{\emph{FeaScatter}})~\cite{zhang2019defense}.
We further compare with \textbf{\emph{TRADES}}~\cite{zhang2019theoretically} in Sec.~\ref{sec:trades}.
\textbf{(E) Robustness Evaluation:}
To evaluate the adversarial robustness of these models, 
%we resort to various attack methods as surrogates. However, a defense model can overfit some attacks, making it hard to fully evaluate the robustness. To overcome this issue, 
we adopt a plenty of attack methods $\mathbf{A}$, and report the \emph{per-example accuracy} as suggested in~\cite{carlini2019evaluating}, which calculates the robust accuracy by 
\begin{equation}\label{eq:12}
%\small
    \mathcal{A}_{\text{rob}} = \frac{1}{n_{\text{test}}}\sum_{i=1}^{n_{\text{test}}}\min_{a\in\mathbf{A}}\mathbb{I}\big(\argmax \{f_{\vect{\theta}}(a(\vect{x}_i))\} = y_i\big),
\end{equation}
where $a(\vect{x}_i)$ is the adversarial example given by attack $a$, and $\mathbb{I}(\cdot)$ is the indicator function. 

\begin{table}[t]
  \caption{Classification accuracy of the three proposed methods and baselines on CIFAR-10 under white-box attacks with $\epsilon=8/255$. The last column shows the overall robustness of the models. We mark the best results for each attack and the overall results that outperform the baselines in \textbf{bold}, and the overall best result in \textcolor{blue}{\textbf{blue}}. We highlight the results of AT\textsubscript{FGSM} and FeaScatter in \textcolor{orange}{orange} to emphasize that these models have the generalization problem across attacks, whose overall robustness is weak.}
  \centering
 \scriptsize
  %\begin{sc}
  \begin{tabular}{c||p{9ex}<{\centering}|p{9ex}<{\centering}|p{9ex}<{\centering}|p{9ex}<{\centering}|p{9ex}<{\centering}|p{9ex}<{\centering}|p{9ex}<{\centering}|p{9ex}<{\centering}}
  \hline
Model & $\mathcal{A}_{\mathrm{nat}}$ & FGSM & PGD-20 & PGD-100 & MIM & C\&W & FeaAttack &  $\mathcal{A}_{\mathrm{rob}}$ \\
\hline\hline
Standard & \bf94.81\% & 12.05\% & 0.00\% & 0.00\% & 0.00\% & 0.00\% & 0.00\% & 0.00\%\\
AT\textsubscript{FGSM} & 93.80\% & \textcolor{orange}{\bf 79.86\%} & 0.12\% & 0.04\% & 0.06\% & 0.13\% & 0.01\% & 0.01\%\\
AT\textsubscript{PGD}$^{\dagger}$ & 87.25\% & 56.04\% & 45.88\% & 45.33\% & 47.15\% & 46.67\% & 46.01\% & 44.89\%\\
AT\textsubscript{PGD} & 86.91\% & 58.30\% & 50.03\% & 49.40\% & 51.40\% & 50.23\% & 50.46\% & 48.26\%\\
ALP & 86.81\% & 56.83\% & 48.97\% & 48.60\% & 50.13\% & 49.10\% & 48.51\% & 47.90\% \\
FeaScatter & 89.98\% & \textcolor{orange}{\bf 77.40\%} & \textcolor{orange}{\bf 70.85\%} & \textcolor{orange}{\bf 68.81\%} & \textcolor{orange}{\bf 72.74\%} & \textcolor{orange}{\bf 58.46\%} & 37.45\% & \textcolor{orange}{37.40\%}\\
\hline
ADT\textsubscript{EXP} & 86.89\% & 60.41\% & 52.18\% & \bf51.69\% & \bf53.27\% & 52.49\% & \bf52.38\% & \textcolor{blue}{\bf50.56\%}\\
ADT\textsubscript{EXP-AM} & 87.82\% & 62.42\% & 51.95\% & 51.26\% & 52.99\% & 51.75\% & 52.04\% & \bf 50.04\%\\
ADT\textsubscript{IMP-AM} & 88.00\% & \bf64.89\% & \bf52.28\% & 51.23\% & 52.64\% & \bf52.65\% & 51.89\% & \bf 49.81\%\\
%\hline\hline
%TRADES & 83.89\% & 59.79\% & 53.92\% & 53.68\% & 54.78\% & 52.71\% & 54.38\% & 51.67\%\\
%\hline
%Unamortized-T & \bf84.89\% & 60.23\% & 54.96\% & 54.56\% & \bf55.72\% & \bf52.96\% & 54.96\% & \bf52.13\%\\
%Gaussian-T & 84.86\% & 60.43\% & 55.22\% & \bf55.00\% & 55.68\% & 52.30\% & 57.91\% & 51.77\%\\
%Implicit-T & 83.95\% & \bf65.61\% & \bf55.87\% & 54.79\% & 54.14\% & 52.82\% & \bf58.80\% & 51.81\%\\
  \hline
   \end{tabular}
 % \end{sc}
  \label{table:cifar10}
  %\vspace{-1ex}
\end{table}

%Refer to Appendix C for the detailed architectures.

%We implement AT\textsubscript{FGSM} and ALP by ourselves and use the pretrained model of FeaScatter.
%Although we use the cross-entropy loss for training, our proposed ADT is fully compatible with other losses. We also integrate the \textbf{TRADES} loss \cite{zhang2019theoretically} with ADT, and provide the results in Appendix D.3.

%\vspace{-1.2ex}
\subsection{Robustness under white-box attacks}\label{sec:white}
%\vspace{-1.2ex}

We first compare the robustness of the proposed methods with baselines under various white-box attacks.
We adopt FGSM~\cite{goodfellow2014explaining}, PGD~\cite{madry2017towards}, MIM~\cite{Dong2017}, C\&W~\cite{carlini2017towards}, and a feature attack (FeaAttack)~\cite{lin2019feature} for evaluation. C\&W is implemented by adopting the margin-based loss function in~\cite{carlini2017towards} and using PGD for optimization. We use $20$ and $100$ steps for PGD, $20$ steps for MIM, and $30$ steps for C\&W. The step size is $\alpha=\nicefrac{\epsilon}{4}$ in these attacks. 
%FeaAttack is a stronger attack for the defense FeaScatter.
%which is robust against first-order adversaries but can be defeated by FeaAttack. 
The details of FeaAttack are provided in Appendix~\ref{sec:c-5}.

\begin{table}[t]
%\vspace{-1ex}
\caption{Classification accuracy of the three proposed methods and baselines on CIFAR-100 and SVHN under part of white-box attacks. Full accuracy results on all adopted white-box attacks can be found in Appendix~\ref{sec:d-1}}\label{table:other}
\begin{subtable}{0.48\linewidth}
  \centering
 \scriptsize
  %\begin{sc}
  \begin{tabular}{c||c|c|c|c}
  \hline
Model & $\mathcal{A}_{\mathrm{nat}}$ & PGD-20 & PGD-100 &  $\mathcal{A}_{\mathrm{rob}}$ \\
\hline\hline
Standard & \bf78.59\% & 0.02\% & 0.01\% & 0.00\%\\
AT\textsubscript{PGD} & 61.45\% & 25.71\% & 25.40\% & 24.49\%\\
\hline
ADT\textsubscript{EXP} & 62.70\% & 28.96\% & \bf28.60\% & \textcolor{blue}{\bf27.13\%}\\
ADT\textsubscript{EXP-AM} & 62.84\% & 29.01\% & 28.46\% & \bf26.87\%\\
ADT\textsubscript{IMP-AM} & 64.07\% & \bf29.40\% & 28.43\% & \bf26.80\%\\
  \hline
  \end{tabular}
  \vspace{-1.5ex}
  \caption{CIFAR-100, $\epsilon=8/255$.}
  \end{subtable}
  \hspace{1ex}
  \begin{subtable}{0.48\linewidth}
\centering
 \scriptsize
  %\begin{sc}
  \begin{tabular}{c||c|c|c|c}
  \hline
Model & $\mathcal{A}_{\mathrm{nat}}$ & PGD-20 & PGD-100 &  $\mathcal{A}_{\mathrm{rob}}$ \\
\hline\hline
Standard & \bf96.12\% & 3.64\% & 2.95\% & 2.14\%\\
AT\textsubscript{PGD} & 95.07\% & 74.22\% & 73.79\% & 73.38\%\\
\hline
ADT\textsubscript{EXP} & 95.70\% & \bf77.01\% & \bf76.62\% & \textcolor{blue}{\bf 75.55\%}\\
ADT\textsubscript{EXP-AM} & 95.67\% & 76.12\% & 75.58\% & \bf75.00\%\\
ADT\textsubscript{IMP-AM} & 95.62\% & 75.61\% & 74.85\% & \bf74.13\%\\
\hline
   \end{tabular}
   \vspace{-1.5ex}
   \caption{SVHN, $\epsilon=4/255$.}
  \end{subtable}
  \vspace{-3ex}
\end{table}

On \textbf{CIFAR-10}, we show the classification accuracy of the proposed methods --- ADT\textsubscript{EXP}, ADT\textsubscript{EXP-AM}, ADT\textsubscript{IMP-AM}, and baseline models --- Standard, AT\textsubscript{FGSM}, AT\textsubscript{PGD}$^{\dagger}$, AT\textsubscript{PGD}, ALP, FeaScatter on natural inputs and adversarial examples in Table~\ref{table:cifar10}. It is obvious that some AT-based methods exhibit the generalization problem across attacks, i.e., AT\textsubscript{FGSM} and FeaScatter, whose overall robustness is weak. But ADT-based methods do not have this issue by showing consistent robustness performance across all tested attacks. Although AT\textsubscript{PGD} does not have this issue also, and achieves the best performance among the AT-based defenses, the proposed ADT reveals improved overall robustness than AT\textsubscript{PGD}, showing the effectiveness. We show the results on \textbf{CIFAR-100} and \textbf{SVHN} in Table~\ref{table:other}. The results consistently demonstrate that ADT-based methods can outperform AT\textsubscript{PGD} under white-box attacks.

It can be further seen that ADT\textsubscript{EXP} is better than ADT\textsubscript{EXP-AM} and ADT\textsubscript{IMP-AM} in most cases. We suspect the reason is that amortizing the adversarial distributions through a generator network is hard to learn appropriate adversarial regions for every input, owing to the limited capacity of the generator. Nevertheless, it can accelerate training, as shown in Appendix~\ref{sec:d-4}.
Note that ADT\textsubscript{IMP-AM} obtains similar robustness with ADT\textsubscript{EXP-AM}. It indicates that though the adopted implicit distributions enable us to optimize in a larger distribution family and the optimization always converges to local optima, ADT\textsubscript{IMP-AM} does not necessarily lead to better adversarial distributions and more robust models.
% it is hard to learn a better implicit adversarial distribution than the explicit one, which may be improved in the future.

\begin{table}[t]
%\vspace{-0.2cm}
\begin{minipage}{0.5\linewidth}
\setlength{\tabcolsep}{5pt}
\caption{Accuracy on CIFAR-10 under SPSA attack with different batch sizes and $\epsilon=8/255$.}
  \centering
 \scriptsize
  \begin{tabular}{c||c|c|c|c}
  \hline
Model & SPSA\textsubscript{256} & SPSA\textsubscript{512} & SPSA\textsubscript{1024} & SPSA\textsubscript{2048} \\
\hline\hline
Standard & 0.00\% & 0.00\% & 0.00\% & 0.00\%\\
AT\textsubscript{PGD} & 60.67\% & 58.10\% & 55.82\% & 54.37\%\\
\hline
ADT\textsubscript{EXP} & 62.22\% & 59.94\% & \bf57.97\% & \bf56.27\% \\
ADT\textsubscript{EXP-AM} & \bf62.58\% & \bf60.12\% & 57.62\% & 55.84\% \\
ADT\textsubscript{IMP-AM} & 62.49\% & 59.77\% & 57.34\% & 55.67\% \\
\hline
   \end{tabular}
  \label{table:spsa}
  \end{minipage}
  \hspace{1ex}
  \begin{minipage}{0.48\linewidth}
  \setlength{\tabcolsep}{5pt}
     \caption{Accuracy on CIFAR-10 under PGD-20, EXP, EXP-AM, and IMP-AM attacks with $\epsilon=8/255$.}
  \centering
  \scriptsize
  %\begin{sc}
  \begin{tabular}{c||c|c|c|c}
  \hline
Model & PGD-20 & EXP & EXP-AM & IMP-AM\\
\hline\hline
Standard & 0.00\% & 0.00\% & 9.24\% & 9.83\%\\
AT\textsubscript{PGD} & 50.03\% & 49.97\% & 50.46\% & 50.36\%\\
\hline
ADT\textsubscript{EXP} & 52.18\% & 51.96\% & 52.71\% & 52.82\%\\
% & 49.93\% & 49.42\%\\
ADT\textsubscript{EXP-AM} & 51.95\% & 51.62\% & 52.85\% & 52.72\%\\
ADT\textsubscript{IMP-AM} & 52.28\% & 51.46\% & 52.76\% & 52.48\%\\
  \hline
   \end{tabular}
  \label{table:g_attack}
  \end{minipage}
  \vspace{-1ex}
\end{table}

\begin{wrapfigure}{r}{0.5\linewidth}
\centering\vspace{-2ex}
\includegraphics[width=1.0\linewidth]{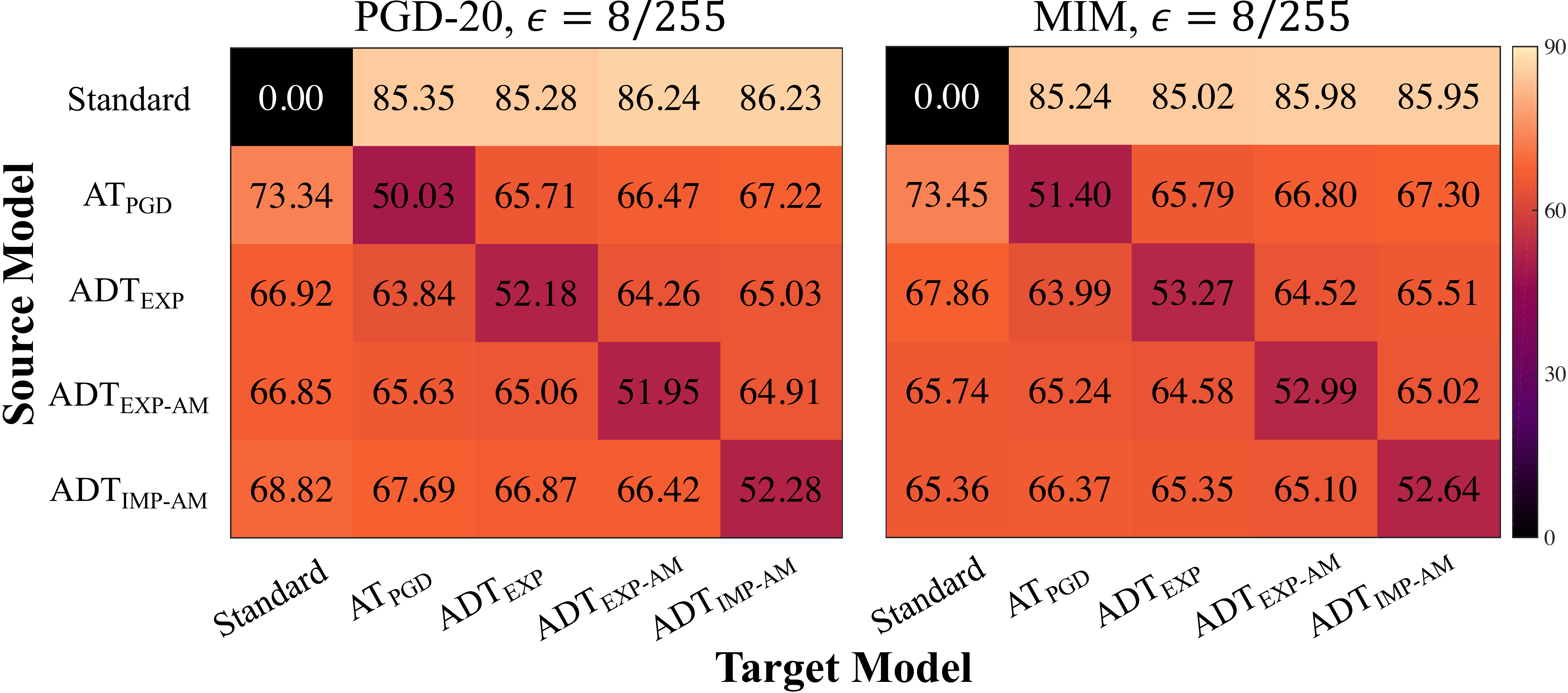}
\vspace{-3.7ex}
\caption{Classification accuracy (\%) under transfer-based black-box attacks. The \emph{source model} refers to the one used to craft adversarial examples, and the \emph{target model} is the one being attacked.}
%\tianyu{use larger font size of the reported values.}}
\vspace{-2ex}
\label{fig:transfer}
\end{wrapfigure}

%\vspace{-1.2ex}
\subsection{Robustness under black-box attacks}\label{sec:black}
%\vspace{-1.2ex}

Now we evaluate the robustness of the defenses on CIFAR-10 under black-box attacks to perform a thorough evaluation~\cite{carlini2019evaluating}. We first evaluate \emph{transfer-based black-box attacks} using PGD-20 and MIM. The results in Fig.~\ref{fig:transfer} show that these models obtain higher accuracy under transfer-based attacks than white-box attacks.
We further perform \emph{query-based black-box attacks} using SPSA~\cite{uesato2018adversarial} and report the results in Table~\ref{table:spsa}. To estimate the gradients, we set the batch size as $256$, $512$, $1024$, and $2048$, the perturbation size as $0.001$, and the learning rate as $0.01$. We run SPSA attacks for $100$ iterations, and early-stop when we cause misclassification.
The accuracy under SPSA is higher than that under white-box attacks. And our methods obtain better robustness over AT\textsubscript{PGD}.
In summary, the black-box results verify that our methods can reliably improve the robustness rather than causing gradient masking~\cite{Athalye2018Obfuscated}.

%\vspace{-1.2ex}
\subsection{Additional results and ablation studies}\label{sec:ablation}
%\vspace{-1.2ex}
%We provide more results to investigate the impacts of several components in ADT.

\textbf{Attack performance of adversarial distributions.}
First, we explore the attack performance of the three proposed methods (i.e., EXP, EXP-AM, and IMP-AM) for learning the adversarial distributions. 
%We choose Standard, AT\textsubscript{PGD}, ADT\textsubscript{EXP}, ADT\textsubscript{EXP-AM}, and ADT\textsubscript{IMP-AM} as the target models. 
For EXP, we set $T=20$, $k=10$ to conduct a more powerful attack. 
We further study the convergence of EXP in Appendix~\ref{sec:d-3}. 
For EXP-AM and IMP-AM, we retrain the generator networks for each pretrained defense. The attack results on five models are shown in Table~\ref{table:g_attack}. From the results, EXP is slightly stronger than PGD-20 while EXP-AM and IMP-AM exhibit comparable attack power.

\begin{wrapfigure}{r}{0.46\linewidth}
\begin{minipage}{.48\linewidth}
\begin{center}
\centerline{\includegraphics[width=1.0\linewidth]{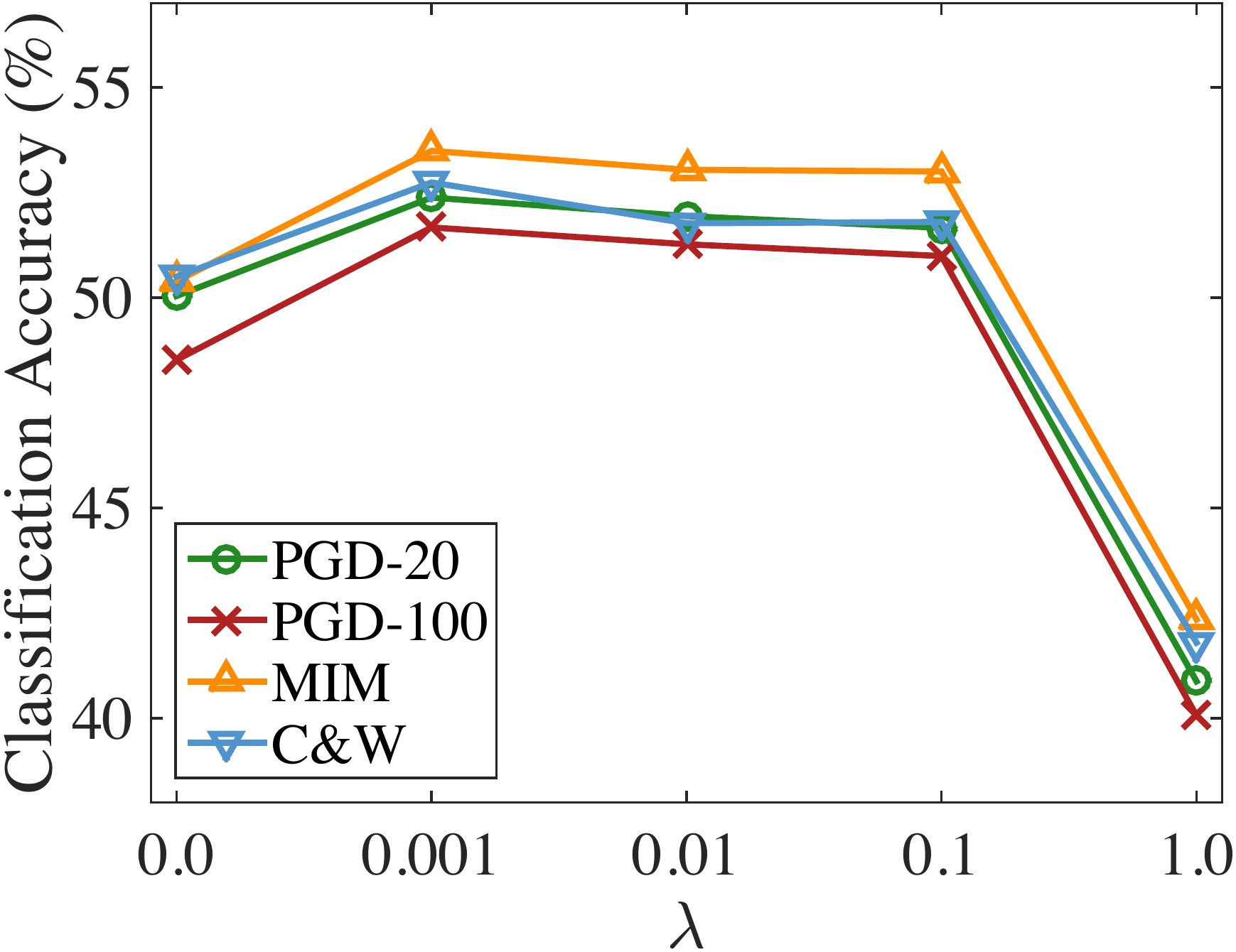}}
\end{center}
\end{minipage}
\begin{minipage}{.48\linewidth}
\begin{center}
\centerline{\includegraphics[width=1.0\linewidth]{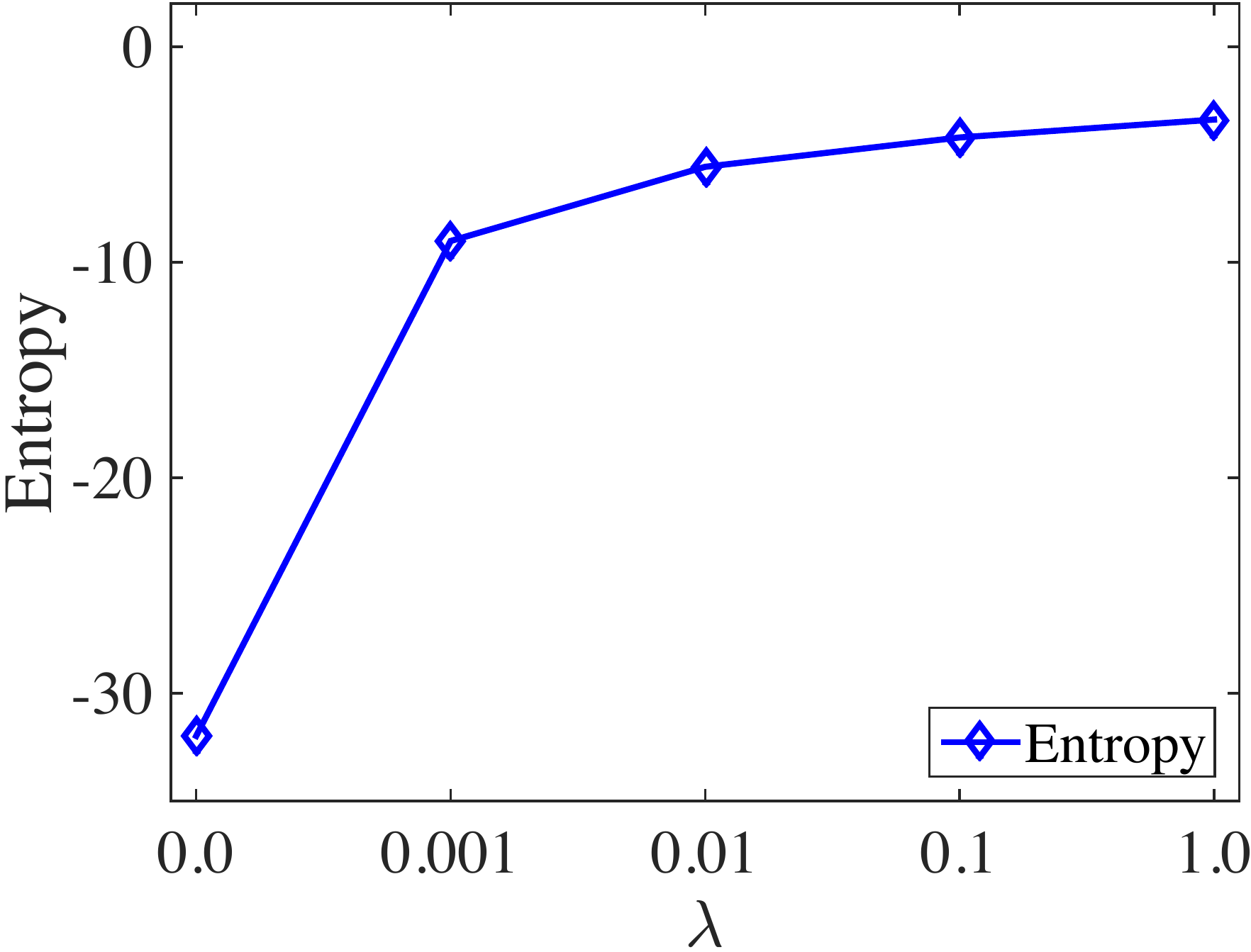}}
\end{center}
\end{minipage}
\vspace{-3.5ex}
\caption{Classification accuracy (\%) under white-box attacks and the average entropy of the adversarial distributions of ADT\textsubscript{EXP-AM} on CIFAR-10 trained with $\lambda=0.0$, $0.001$, $0.01$, $0.1$, and $1.0$.}
\vspace{-2ex}
\label{fig:lambda}
\end{wrapfigure}

\textbf{The impact of $\lambda$.}
We study the impact of $\lambda$ on the performance of ADT. We choose ADT\textsubscript{EXP-AM} as a case study for its fast training process and analytical entropy estimation. Fig.~\ref{fig:lambda} shows the robustness under white-box attacks and the average entropy of the adversarial distributions of ADT\textsubscript{EXP-AM} trained with $\lambda=0.0$, $0.001$, $0.01$, $0.1$, and $1.0$. Generally, a larger $\lambda$ leads to a larger entropy and better robustness. But a too large $\lambda$ will reduce the robustness.

\textbf{Loss landscape analysis.}
%We conduct an additional check against gradient masking~\cite{Athalye2018Obfuscated} by looking at the loss landscape. 
First, we plot the cross-entropy loss of the models projected along the gradient direction ($\vect{d_g}$) and a random direction ($\vect{d_r}$) in the vicinity of a natural input in Fig.~\ref{fig:loss}. 
%whose smoothness may highly correlate with the adversarial robustness.
%considering that the local smoothness of the loss surface around the natural input is highly correlated with the adversarial robustness. 
%As we can see, the adversarially trained models have more smooth loss surface than the commonly trained model (i.e., Standard), validating the effectiveness of adversarial training. Besides, the models trained by ADT exhibit smoother and flatter loss surface than the model trained by AT. In other words, ADT manages to force the points in the vicinity of a input to have lower loss, thus it is harder to create adversarial examples for attack.
Notably, the models trained by ADT exhibit smoother and more flattened loss surfaces than Standard and AT\textsubscript{PGD}, and thus deliver better robustness.
%In the case of a non-smooth loss surface around the input, it is relatively easy to find some adversarial region which contains points with high loss; while in the case of a smooth and flat loss surface,  
We further quantitatively measure the smoothness of loss surfaces with the dominant eigenvalue of the Hessian matrix of the classification loss w.r.t. the input as a proxy.
%for the smoothness at the natural input. 
We use $1000$ images from the test set of CIFAR-10 for calculation, and report the mean and standard derivation in Fig.~\ref{fig:loss}(f). The numbers are consistent with the visualization results and help us confirm the superiority of ADT upon AT to learn smooth loss surfaces and robust deep models.

%\vspace{-1.2ex}
\subsection{Compare with the state-of-the-art}
\label{sec:trades}
%\vspace{-1.2ex}
%We conduct two more experiments to compare the performance of ADT with the state-of-the-art results.
Although we use the cross-entropy loss as our training objective in previous experiments, our proposed ADT framework is compatible with other loss functions. In this section, we integrate TRADES~\cite{zhang2019theoretically}, a state-of-the-art AT method, with ADT.
We implement ADT by using the TRADES loss in Eq.~\eqref{eq:adt-ent}. We follow the same experimental settings as in~\cite{zhang2019theoretically}, where a WRN-34-10 model is used and $\epsilon$ is $0.031$. We evaluate the robustness by all adopted attacks. We show part of the results in Table~\ref{table:trades}, and leave full results in Appendix~\ref{sec:d-2}, which prove that the proposed methods also outperform TRADES.

\begin{table}[t]
\caption{Comparison of ADT with TRADES on CIFAR-10. $\beta$ is a hyperparameter balancing the trade-off between natural and robust accuracy. The results of TRADES are reproduced based on the official open-sourced code of~\cite{zhang2019theoretically}.}
  \centering
 \scriptsize
 \setlength{\tabcolsep}{4.5pt}
  \begin{tabular}{c|c|c|c|c|c||c|c|c|c|c|c}
  \hline
Model & $\beta$ & $\mathcal{A}_{\mathrm{nat}}$ & PGD-20 & PGD-100 & $\mathcal{A}_{\mathrm{rob}}$ & Model & $\beta$ & $\mathcal{A}_{\mathrm{nat}}$ & PGD-20 & PGD-100 & $\mathcal{A}_{\mathrm{rob}}$ \\
\hline\hline
TRADES & 1.0 & 87.99\% & 51.08\% & 48.41\% & 47.75\% & TRADES & 6.0 & 84.02\% & 56.06\% & 54.49\% & 52.64\% \\
ADT\textsubscript{EXP} & 1.0 & \bf89.74\% & 52.39\% & 49.88\% & \bf49.05\% & ADT\textsubscript{EXP} & 6.0 & 84.66\% & 57.71\% & \bf56.17\% & \textcolor{blue}{\bf 54.21\%}\\
ADT\textsubscript{EXP-AM} & 1.0 & 88.86\% & \bf54.44\% & \bf51.66\% & \textcolor{blue}{\bf50.78\%} & ADT\textsubscript{EXP-AM} & 6.0 & 84.85\% & 57.67\% & 55.73\% & \bf54.09\% \\
ADT\textsubscript{IMP-AM} & 1.0 & 88.80\% & 54.22\% & 51.09\% & \bf50.14\% & ADT\textsubscript{IMP-AM} & 6.0 & \bf84.96\% & \bf57.82\% & 55.45\% & \bf53.66\%\\
\hline
   \end{tabular}
  \label{table:trades}
  %\vspace{-2ex}
\end{table}

\begin{wraptable}{r}{68mm}
\vspace{-2.5ex}
  \caption{Comparison with the state-of-the-art AT\textsubscript{PGD} model in~\cite{rice2020overfitting}.}
  \vspace{-1.5ex}
  \label{tab:1}
  \centering
  \scriptsize
  \begin{tabular}{c||c|c|c|c}
    \hline
    Model & $\mathcal{A}_{\mathrm{nat}}$ & PGD-10 & PGD-20 & PGD-100\\
    \hline
    AT\textsubscript{PGD} & 86.41\% & 55.90\% & 54.52\% & 54.20\% \\
    ADT\textsubscript{EXP} & 86.49\% & \bf56.84\% & \bf55.43\% & \bf55.01\% \\
    ADT\textsubscript{EXP-AM} & 87.27\% & 56.28\% & 54.88\% & 54.58\% \\
    ADT\textsubscript{IMP-AM} & \bf87.38\% & 56.63\% & 55.10\% & 54.43\% \\
    \hline
  \end{tabular}\vspace{-2ex}
\end{wraptable}
Besides, a recent state-of-the-art AT\textsubscript{PGD} model is obtained in~\cite{rice2020overfitting}. It achieves better robustness by using early stopping and a proper weight decay value. To fairly compare with this model, we reproduce the results of~\cite{rice2020overfitting} and train ADT based models using the same settings/hyperparameters as in~\cite{rice2020overfitting}, where a WRN-34-10 model is adopted.
The results of those models on CIFAR-10 are shown in Table~\ref{tab:1}. By using the same training settings, our models can also improve the performance over AT\textsubscript{PGD}.

%\vspace{-1.5ex}
\section{Conclusion}
%\vspace{-1.2ex}
In this paper, we introduced an adversarial distributional training framework for learning robust DNNs.
ADT can learn an adversarial distribution to characterize heterogeneous adversarial examples around a natural one under an entropic regularizer.
Through a theoretical analysis, we provided a general algorithm for solving ADT, and proposed to parameterize the adversarial distributions in ADT with three different approaches, ranging from the typical Gaussian distributions to the flexible implicit distributions.
We conducted extensive experiments on CIFAR-10, CIFAR-100, and SVHN to demonstrate the effectiveness of ADT on building robust DNNs, compared with the state-of-the-art adversarial training methods. 

\section*{Broader Impact}

The existence of adversarial examples poses potential security threats to machine learning models, when they are deployed to real-world applications, especially the security-sensitive ones, such as autonomous driving, healthcare, and finance.
The model vulnerability to such small perturbations could lower the confidence of the public on machine learning techniques.
Therefore, it is of particular importance to develop more robust models. This work is dedicated to developing a new learning framework to train robust deep learning models, which is the potential positive impact of this work in the society.
Nevertheless, many works have shown that there is an inherent trade-off between robustness and natural accuracy~\cite{tsipras2019robustness,zhang2019theoretically}, that a classifier trained to be adversarially robust would introduce degraded accuracy on clean data, and our work is no exception. Although our proposed methods can obtain higher natural accuracy than the previous adversarial training methods, they still have lower natural accuracy than a standard trained model. 
The degeneration in natural accuracy could be a negative consequence.
From a different perspective, adversarial examples also provide an opportunity to protect private information of users~\cite{oh2017adversarial,yang2020towards}. Building a robust model could negatively impact users' ability to hide their privacy from the excessive unauthorized recognition systems.

\section*{Acknowledgements}
This work was supported by the National Key Research and Development Program of China (No.2017YFA0700904), NSFC Projects (Nos. 61620106010, U19B2034, U1811461), Beijing
Academy of Artificial Intelligence (BAAI), Tsinghua-Huawei Joint Research Program, a grant from Tsinghua Institute for Guo Qiang, Tiangong Institute for Intelligent Computing, and the NVIDIA NVAIL Program with GPU/DGX Acceleration.
Yinpeng Dong was supported by MSRA and Baidu fellowships.

{\small
\bibliographystyle{plainnat}
\bibliography{reference}}

\clearpage
\appendix
\numberwithin{equation}{section}

\section{Technical details and algorithms} \label{sec:a}

\subsection{ADT\textsubscript{EXP}}
We provide the algorithm for ADT\textsubscript{EXP} in Alg.~\ref{algo-1}.

\begin{algorithm}[h]
   \caption{The training algorithm for ADT\textsubscript{EXP}}
   \label{algo-1}
\begin{algorithmic}[1]
   \Require Training data $\mathcal{D}$, objective function $\mathcal{J}\big(p_{\vect{\phi}_i}(\vect{\delta}_i), \vect{\theta}\big)$, training epochs $N$, the number of inner maximization steps $T$, the number of MC samples for gradient estimation in each step $k$,  and learning rates $\eta_{\vect{\theta}}$, $\eta_{\vect{\phi}}$.
   \State  Initialize $\vect{\theta}$;
   \For {$\text{epoch}=1$ {\bfseries to} $N$}
   \For {each minibatch $\mathcal{B}\subset\mathcal{D}$}
   \For {each input $(\vect{x}_i, y_i) \in \mathcal{B}$}
   \State Initialize $\vect{\phi}_i$;
   \For{$t = 1$ {\bfseries to} $T$}
   \State Calculate the gradient $\vect{g}_i$ of $\vect{\phi}_i$ by Eq.~\eqref{eq:adt_gau_der} via MC integration using $k$ samples;
   \State Update $\vect{\phi}_i$ with gradient ascent
       \[\vect{\phi}_i\leftarrow\vect{\phi}_i+\eta_{\vect{\phi}}\cdot\vect{g}_i.\]
    \EndFor
   \EndFor
       
    \State Update $\vect{\theta}$ with stochastic gradient descent
       \[\vect{\theta}\leftarrow\vect{\theta}-\eta_{\vect{\theta}}\cdot\mathbb{E}_{(\vect{x}_i, y_i)\in\mathcal{B}}\big[\nabla_{\vect{\theta}}\mathcal{J}\big(p_{\vect{\phi}_i}(\vect{\delta}_i), \vect{\theta}\big)\big].\]
   \EndFor
   \EndFor
\end{algorithmic}
\end{algorithm}

\subsection{ADT\textsubscript{EXP-AM}}

By amortizing the explicit adversarial distributions, we can rewrite the minimax problem of ADT as 
\begin{equation}
\label{eq:adt-ent-explicit}
    \min_{\vect{\theta}}\max_{\vect{\phi}}\frac{1}{n}\sum_{i=1}^{n}\Big\{\mathbb{E}_{p_{\vect{\phi}}(\vect{\delta}_i|\vect{x}_i)}\big[ \mathcal{L}(f_{\vect{\theta}}(\vect{x}_i + \vect{\delta}_i), y_i)\big] + \lambda\mathcal{H}(p_{\vect{\phi}}(\vect{\delta}_i|\vect{x}_i))\Big\}, 
\end{equation}
where $\vect{\theta}$ and $\vect{\phi}$ are the parameters of the DNN classifier and the generator, respectively.
%As we stated, the generator is optimized to represent a generalizable mapping, thus we reasonably hypothesize that it can produce relatively adversarial distributions for the new incoming training examples. Therefore, 
During training, we perform stochastic gradient descent and ascent on $\vect{\theta}$ and $\vect{\phi}$ simultaneously, to accomplish adversarial training. To enable the gradients flowing from $\vect{\delta}_i$ to $\vect{\phi}$, we apply the same reparameterization strategy as in Sec.~\ref{sec:adt_gau}. In practice, we only use one MC sample for each data. We provide the algorithm for ADT\textsubscript{EXP-AM} in Alg.~\ref{algo-2}.

\begin{algorithm}[h]
   \caption{The training algorithm for ADT\textsubscript{EXP-AM}}
   \label{algo-2}
\begin{algorithmic}[1]
   \Require Training data $\mathcal{D}$, objective function in Eq.~\eqref{eq:adt-ent-explicit}, training epochs $N$, and learning rates $\eta_{\vect{\theta}}$, $\eta_{\vect{\phi}}$.
   \State Initialize $\vect{\theta}$ and $\vect{\phi}$;
   \For{$\text{epoch}=1$ {\bfseries to} $N$}
   \For{each minibatch $\mathcal{B}\subset\mathcal{D}$}
   \State Input $\vect{x}_i$ to the generator and obtain the distribution parameters $(\vect{\mu}_i, \vect{\sigma}_i)$ for each data $(\vect{x}_i, y_i) \in \mathcal{B}$;
   \State Sample one $\vect{\delta}_i$ from the distribution defined by Eq.~\eqref{eq:explicit} given $(\vect{\mu}_i, \vect{\sigma}_i)$ for each $(\vect{x}_i, y_i) \in \mathcal{B}$ to approximately calculate the gradient of Eq.~\eqref{eq:adt-ent-explicit} w.r.t. $\vect{\theta}$ and $\vect{\phi}$, and obtain $\vect{g}_{\vect{\theta}}$ and $\vect{g}_{\vect{\phi}}$;
    \State Update $\vect{\theta}$ by: $\vect{\theta}\leftarrow\vect{\theta}-\eta_{\vect{\theta}}\cdot\vect{g}_{\vect{\theta}}$.
    \State Update $\vect{\phi}$ by: $\vect{\phi}\leftarrow\vect{\phi}+\eta_{\vect{\phi}}\cdot\vect{g}_{\vect{\phi}}$.
   \EndFor
   \EndFor
\end{algorithmic}
\end{algorithm}

%This approach takes a set of shared parameters as a substitute for individual distribution parameters, to amortize the optimization over all $\vect{x}_i \in \mathcal{D}$. When given a natural example $\vect{x}_i$, we can not only simply utilize the explicit information like loss and gradient pertain to $\vect{x}_i$, but also leverage the implicit wisdom of all previous examples we have ever trained on, to produce improved adversarial distribution. \yinpeng{What's the meaning?} 
\subsection{ADT\textsubscript{IMP-AM}}

For the implicit adversarial distributions, we have no access to the density $p_{\vect{\phi}}(\vect{\delta}_i|\vect{x}_i)$, such that the entropy of the adversarial distributions cannot be estimated exactly\footnote{We can also directly estimate the gradient of the entropy with advanced techniques such as spectral Stein gradient estimator~\cite{shi2018spectral}, and we leave this for future work.}. An appealing alternative is to maximize the variational lower bound of the entropy~\cite{dai2017calibrating} for its simplicity and success in GANs~\cite{NIPS2017_7229}.
In our case, for a natural input $\vect{x}_i$, we can similarly derive the following lower bound stemming from the mutual information between the perturbation $\vect{\delta}_i$ and the random noise $\vect{z}$ (proof in Appendix~\ref{sec:b-3}) as
\begin{equation}
\label{eq:ent-var}
    \mathcal{H}(p_{\vect{\phi}}(\vect{\delta}_i|\vect{x}_i)) \geq \mathcal{U}(q) = \mathbb{E}_{p(\vect{z})}\log q(\vect{z} | g_{\vect{\phi}}(\vect{z};\vect{x}_i)) + c,
\end{equation}
%\yinpeng{When use $z\sim p(z)$ or $p(z)$?}
where $c$ is a constant and $q(\cdot|\cdot)$ is an introduced variational distribution. 
%Maximizing $\mathcal{U}(q)$ can effectively maximize the entropy term $\mathcal{H}(p_{\vect{\phi}}(\vect{\delta}_i|\vect{x}_i))$. 
In practice, we implement $q$ as a diagonal Gaussian, whose mean and standard derivation are given by a $\vect{\psi}$-parameterized neural network. Then we have the training objective as
\begin{equation}
\label{eq:adt-ent-implicit-final}
    \min_{\vect{\theta}}\max_{\vect{\phi},\vect{\psi}}\frac{1}{n}\sum_{i=1}^{n}\Big\{\mathbb{E}_{p(\vect{z})}\big[ \mathcal{L}(f_{\vect{\theta}}(\vect{x}_i + g_{\vect{\phi}}(\vect{z};\vect{x}_i)), y_i) + \lambda\log q_{\vect{\psi}}(\vect{z} | g_{\vect{\phi}}(\vect{z};\vect{x}_i))\big]\Big\},
\end{equation}
which is solved by simultaneous stochastic gradient descent and ascent on $\vect{\theta}$ and $(\vect{\phi}, \vect{\psi})$.
We provide the algorithm for ADT\textsubscript{IMP-AM} in Alg.~\ref{algo-3}.

\begin{algorithm}[h]
   \caption{The training algorithm for ADT\textsubscript{IMP-AM}}
   \label{algo-3}
\begin{algorithmic}[1]
   \Require Training data $\mathcal{D}$, objective function in Eq.~\eqref{eq:adt-ent-implicit-final}, training epochs $N$, and learning rates $\eta_{\vect{\theta}}$, $\eta_{\vect{\phi}}$, $\eta_{\vect{\psi}}$.
   \State Initialize $\vect{\theta}$, $\vect{\phi}$, and $\vect{\psi}$;
   \For{$\text{epoch}=1$ {\bfseries to} $N$}
   \For{each minibatch $\mathcal{B}\subset\mathcal{D}$}
   \State For each $(\vect{x}_i, y_i) \in \mathcal{B}$, sample a noise $\vect{z}_i$ from $\mathrm{U}(-1,1)$.
   \State Use the sampled noises to approximately calculate the gradient of Eq.~\eqref{eq:adt-ent-implicit-final} w.r.t. $\vect{\theta}$, $\vect{\phi}$, and $\vect{\psi}$, and obtain $\vect{g}_{\vect{\theta}}$, $\vect{g}_{\vect{\phi}}$, and $\vect{g}_{\vect{\psi}}$.
    \State Update $\vect{\theta}$ by: $\vect{\theta}\leftarrow\vect{\theta}-\eta_{\vect{\theta}}\cdot\vect{g}_{\vect{\theta}}$.
    \State Update $\vect{\phi}$ by: $\vect{\phi}\leftarrow\vect{\phi}+\eta_{\vect{\phi}}\cdot\vect{g}_{\vect{\phi}}$.
    \State Update $\vect{\psi}$ by: $\vect{\psi}\leftarrow\vect{\psi}+\eta_{\vect{\psi}}\cdot\vect{g}_{\vect{\psi}}$.
   \EndFor
   \EndFor
\end{algorithmic}
\end{algorithm}

\section{Proofs} 
We provide the proofs in this section.
\subsection{Proof of Theorem~\ref{the:1}} \label{sec:b-1}

%\begin{assumption}\label{ass:1}
%The loss function $\mathcal{J}\big(p(\vect{\delta}_i), \vect{\theta}\big)$ is continuously differentiable w.r.t. $\vect{\theta}$.
%\end{assumption}

%\begin{assumption}\label{ass:2}
%Probability density functions of distributions in $\mathcal{P}$ are bounded and equicontinuous.
%\end{assumption}

%\begin{theorem}\label{the:1}
%Suppose Assumptions~\ref{ass:1} and \ref{ass:2} hold. We define $\rho(\vect{\theta})=\max_{p(\vect{\delta}_i)\in\mathcal{P}}\mathcal{J}\big(p(\vect{\delta}_i), \vect{\theta}\big)$, and $\mathcal{P}^*(\vect{\theta}) = \{p(\vect{\delta}_i)\in\mathcal{P}:\mathcal{J}\big(p(\vect{\delta}_i), \vect{\theta}\big)=\rho(\vect{\theta})\}$. Then $\rho(\vect{\theta})$ is directionally differentiable, and its directional derivative along the direction $\vect{v}$ satisfies
%\begin{equation}
%    \rho'(\vect{\theta};\vect{v}) = \sup_{p(\vect{\delta}_i) \in \mathcal{P}^*(\vect{\theta})}\vect{v}^\top\nabla_{\vect{\theta}}\mathcal{J}\big(p(\vect{\delta}_i), \vect{\theta}\big).
%\end{equation}
%Particularly, when $\mathcal{P}^*(\vect{\theta}) = \{p^*(\vect{\delta}_i)\}$ only contains one maximizer, $\rho(\vect{\theta})$ is differentiable at $\vect{\theta}$ and
%\begin{equation}
%    \nabla_{\vect{\theta}}\rho(\vect{\theta}) = \nabla_{\vect{\theta}}\mathcal{J}\big(p^*(\vect{\delta}_i), \vect{\theta}\big).
%\end{equation}
%\end{theorem}

\begin{proof}
Recall that $\mathcal{P}$ is a set of distributions, which can be expressed by their probability density functions. The support of these functions is contained in $\mathcal{S}$ and these functions are equicontinuous by Assumption~\ref{ass:2}.
$\mathcal{S} = \{{\vect{\delta}}: \|{\vect{\delta}}\|_{\infty}\leq\epsilon\}$ is the allowed perturbation set. The Euclidean distance $\ell_2$ defines a metric on $\mathcal{S}$.
We let
\begin{equation*}
    \mathcal{C}(\mathcal{S},\mathbb{R}) = \{h: \mathcal{S}\rightarrow\mathbb{R}|h \text{ is continuous}\}
\end{equation*}
be the collection of all continuous functions from $\mathcal{S}$ to $\mathbb{R}$. Then $\mathcal{P}$ is a subset of $\mathcal{C}(\mathcal{S},\mathbb{R})$.
We let
\begin{equation*}
    d_{\mathcal{C}}(p,q) = \max_{\vect{\delta}\in\mathcal{S}}|p(\vect{\delta}) - q(\vect{\delta})|
\end{equation*}
for all $p,q\in\mathcal{C}(\mathcal{S},\mathbb{R})$ be a metric on $\mathcal{C}(\mathcal{S},\mathbb{R})$.
Then we can see that $(\mathcal{C}(\mathcal{S},\mathbb{R}), d_{\mathcal{C}})$ is a metric space.

We state the following lemma to prove that $\mathcal{P}$ is compact.
\begin{lemma}\label{lemma:1}
\textbf{(Arzel\`a-Ascoli's Theorem)} Let $(X, d_X)$ be a compact metric space. A subset $\mathcal{K}$ of $\mathcal{C}(X,\mathbb{R})$ is compact if and only if it is closed, bounded, and equicontinuous.
\end{lemma}

Since $(\mathcal{S}, \ell_2)$ is a compact metric space, and $\mathcal{P}$ is closed, bounded, and equicontinuous given by Assumption~\ref{ass:2}, we can see that $\mathcal{P}$ is compact by Lemma~\ref{lemma:1}.

We next need to prove that the loss function $\mathcal{J}\big(p(\vect{\delta}_i), \vect{\theta}\big)$ is continuously differentiable w.r.t. both $p(\vect{\delta}_i)$ and $\vect{\theta}$, i.e., the gradient $\nabla_{\vect{\theta}}\mathcal{J}\big(p(\vect{\delta}_i), \vect{\theta}\big)$ is joint continuous on $\mathcal{P}\times\mathbb{R}^m$, where $m$ is the dimension of $\theta$.

To prove it, we first define a new metric on $\mathcal{P}\times\mathbb{R}^m$ as
\begin{equation*}
    d_{mix}((p_1,\vect{\theta}_1),(p_2, \vect{\theta}_2)) = d_{\mathcal{C}}(p_1, p_2) + \ell_2(\vect{\theta}_1, \vect{\theta}_2).
\end{equation*}
Then $(\mathcal{P}\times\mathbb{R}^m, d_{mix})$ is a new metric space. 

By definition, given a point $(p_0, \vect{\theta}_0) \in \mathcal{P}\times\mathbb{R}^m$, if for each $\tau>0$, there is a $\gamma>0$, such that \[\ell_2\big(\nabla_{\vect{\theta}}\mathcal{J}\big(p(\vect{\delta}_i), \vect{\theta}\big), \nabla_{\vect{\theta}}\mathcal{J}\big(p_0(\vect{\delta}_i), \vect{\theta}_0\big)\big)<\tau\] whenever $d_{mix}((p, \vect{\theta}), (p_0, \vect{\theta}_0)) < \gamma$, then the function $\nabla_{\vect{\theta}}\mathcal{J}\big(p(\vect{\delta}_i), \vect{\theta}\big)$ is continuous at $(p_0, \vect{\theta}_0)$.
If for all points in $\mathcal{P}\times\mathbb{R}^m$, the function is continuous, then $\nabla_{\vect{\theta}}\mathcal{J}\big(p(\vect{\delta}_i), \vect{\theta}\big)$ is continuous on $\mathcal{P}\times\mathbb{R}^m$.

To show that, we first have
\begin{equation}\label{eq:con}
\begin{split}
&\ell_2\big(\nabla_{\vect{\theta}}\mathcal{J}\big(p(\vect{\delta}_i), \vect{\theta}\big), \nabla_{\vect{\theta}}\mathcal{J}\big(p_0(\vect{\delta}_i), \vect{\theta}_0\big)\big)\\
\leq\; &\ell_2\big(\nabla_{\vect{\theta}}\mathcal{J}\big(p(\vect{\delta}_i), \vect{\theta}\big), \nabla_{\vect{\theta}}\mathcal{J}\big(p(\vect{\delta}_i), \vect{\theta}_0\big)\big)
+  \ell_2\big(\nabla_{\vect{\theta}}\mathcal{J}\big(p(\vect{\delta}_i), \vect{\theta}_0\big), \nabla_{\vect{\theta}}\mathcal{J}\big(p_0(\vect{\delta}_i), \vect{\theta}_0\big)\big).
\end{split}
\end{equation}
We already have that the loss function $\mathcal{J}\big(p(\vect{\delta}_i), \vect{\theta}\big)$ is continuously differentiable w.r.t. $\vect{\theta}$ by Assumption~\ref{ass:1}.
Then given $\frac{\tau}{2}$, there is a $\gamma_1$, such that
\[\ell_2\big(\nabla_{\vect{\theta}}\mathcal{J}\big(p(\vect{\delta}_i), \vect{\theta}\big), \nabla_{\vect{\theta}}\mathcal{J}\big(p(\vect{\delta}_i), \vect{\theta}_0\big)\big) < \frac{\tau}{2}\]
whenever $\ell_2(\vect{\theta}, \vect{\theta}_0) < \gamma_1$.

For the second term of the RHS of Eq.~\eqref{eq:con}, we have
\begin{equation*}
    \begin{split}
        \ell_2\big(\nabla_{\vect{\theta}}\mathcal{J}\big(p(\vect{\delta}_i), \vect{\theta}_0\big), \nabla_{\vect{\theta}}\mathcal{J}\big(p_0(\vect{\delta}_i), \vect{\theta}_0\big)\big)
        =\; & \big\|\nabla_{\vect{\theta}}\big(\mathcal{J}\big(p(\vect{\delta}_i), \vect{\theta}_0\big) - \mathcal{J}\big(p_0(\vect{\delta}_i), \vect{\theta}_0\big)\big)\big\|_2 \\
        =\; & \big\|\int_{\mathcal{S}} \big(p(\vect{\delta}_i) - p_0(\vect{\delta}_i)\big)\nabla_{\vect{\theta}}\mathcal{L}(f_{\vect{\theta}}(\vect{x}_i + \vect{\delta}_i), y_i)d\vect{\delta}_i \big\|_2 \\
        \leq\; & d_{\mathcal{C}}(p, p_0)\cdot\int_{\mathcal{S}}\big\|\nabla_{\vect{\theta}}\mathcal{L}(f_{\vect{\theta}}(\vect{x}_i + \vect{\delta}_i), y_i)\big\|_2d\vect{\delta}_i.
    \end{split}
\end{equation*}
Therefore, for the given $\frac{\tau}{2}$, there is also a $\gamma_2$ which equals to
\[\gamma_2 = \frac{\tau}{2\int_{\mathcal{S}}\big\|\nabla_{\vect{\theta}}\mathcal{L}(f_{\vect{\theta}}(\vect{x}_i + \vect{\delta}_i), y_i)\big\|_2d\vect{\delta}_i},\]
such that
\[\ell_2\big(\nabla_{\vect{\theta}}\mathcal{J}\big(p(\vect{\delta}_i), \vect{\theta}_0\big), \nabla_{\vect{\theta}}\mathcal{J}\big(p_0(\vect{\delta}_i), \vect{\theta}_0\big)\big) < \frac{\tau}{2}\]
whenever $d_{\mathcal{C}}(p, p_0) < \gamma_2$.

Combining the results, for a given $\tau>0$, we can set $\gamma=\gamma_1+\gamma_2$, such that
\[\ell_2\big(\nabla_{\vect{\theta}}\mathcal{J}\big(p(\vect{\delta}_i), \vect{\theta}\big), \nabla_{\vect{\theta}}\mathcal{J}\big(p_0(\vect{\delta}_i), \vect{\theta}_0\big)\big) < \tau\]
whenever $d_{mix}((p, \vect{\theta}), (p_0, \vect{\theta}_0)) < \gamma$. Thus we have proven that the loss function $\mathcal{J}\big(p(\vect{\delta}_i), \vect{\theta}\big)$ is continuously differentiable w.r.t. both $p(\vect{\delta}_i)$ and $\vect{\theta}$.

Given the above results, we can directly apply Danskin's theorem \cite{danskin2012theory} to prove Theorem~\ref{the:1}. We state the Danskin's theorem in the following lemma.

\begin{lemma}\label{lemma:2}
\textbf{(Danskin's Theorem)} Let $\mathcal{Q}$ be a nonempty compact topological space and $h:\mathcal{Q}\times\mathbb{R}^m\rightarrow \mathbb{R}$ be a function satisfying that $h(q,\cdot)$ is differentiable for every $q\in\mathcal{Q}$ and $\nabla_{\vect{\theta}}h(q, \vect{\theta})$ is continuous on $\mathcal{Q}\times\mathbb{R}^m$.
We define $\Psi(\vect{\theta})=\max_{q\in\mathcal{Q}}h(q, \vect{\theta})$, and $\mathcal{Q}^*(\vect{\theta}) = \{q\in\mathcal{Q}:h(q, \vect{\theta})=\Psi(\vect{\theta})\}$. Then $\Psi(\vect{\theta})$ is directionally differentiable, and its directional derivative along the direction $\vect{v}$ satisfies
\begin{equation*}
    \Psi'(\vect{\theta};\vect{v}) = \sup_{q \in \mathcal{Q}^*(\vect{\theta})}\vect{v}^\top\nabla_{\vect{\theta}}h(q, \vect{\theta}).
\end{equation*}
Particularly, when $\mathcal{Q}^*(\vect{\theta}) = \{q^*\}$ only contains one maximizer, $\Psi(\vect{\theta})$ is differentiable at $\vect{\theta}$ and
\begin{equation*}
    \nabla_{\vect{\theta}}\Psi(\vect{\theta}) = \nabla_{\vect{\theta}}h(q^*, \vect{\theta}).
\end{equation*}
\end{lemma}

If we let $\mathcal{Q}=\mathcal{P}$ and $h=\mathcal{J}$ in Lemma~\ref{lemma:2}, we can directly prove Theorem~\ref{the:1}.
\end{proof}

\begin{remark}
For the explicit adversarial distributions defined in Eq.~\eqref{eq:explicit}, we can assume that the mean and standard deviation of each dimension satisfy $|\vect{\mu}_i^{(j)}| < \kappa_{\mu}$ and $\kappa_{\sigma}^{lo} < \vect{\sigma}_i^{(j)} < \kappa_{\sigma}^{up}$, where $\kappa_{\mu}$, $\kappa_{\sigma}^{lo}$, and $\kappa_{\sigma}^{up}$ are constants. Note that they can be easily satisfied since we add an entropic regularization term into the training objective~\eqref{eq:adt-ent}, such that the mean cannot be too large while the standard deviation cannot be too small or too large given Eq.~\eqref{eq:11}. In practice, we can clip $\vect{\mu}_i^{(j)}$ and $\vect{\sigma}_i^{(j)}$ if they are out of the thresholds. Then we can prove that the density functions of the explicit adversarial distributions defined in Eq.~\eqref{eq:explicit} are bounded and equicontinuous, satisfying Assumption~\ref{ass:2}. However, for the implicit adversarial distributions introduced in Sec.~\ref{sec:3-3}, we cannot prove that Assumption~\ref{ass:2} is satisfied. Though unsatisfied, the experiments suggest that we can still rely on Theorem~\ref{the:1} and the general algorithm for training.
\end{remark}
\begin{proof}
Due to the diagonal covariance matrix, each dimension of $p_{\vect{\phi}_i}(\vect{\delta}_i)$ is independent. Thus we only consider one dimension of $\vect{\delta}_i$.
For clarity, we denote $\vect{\mu}_i^{(j)}$, $\vect{\sigma}_i^{(j)}$, $\vect{r}^{(j)}$, $\vect{u}_i^{(j)}$, and $\vect{\delta}_i^{(j)}$ as $\mu$, $\sigma$, $r$, $u$, and $\delta$, respectively.
The probability density function of $\delta$ is (see Appendix~\ref{sec:b-2} for details)
\begin{equation*}
\begin{aligned}
    p(\delta) = & \frac{1}{\sqrt{2\pi}\sigma}\exp{\Big(-\frac{(\frac{1}{2}\log\frac{\epsilon+\delta}{\epsilon-\delta}-\mu)^2}{2\sigma^2}\Big)} \cdot \frac{\epsilon}{\epsilon^2-\delta^2} \\ = & \frac{1}{\sqrt{2\pi}\sigma}\exp{\Big(-\frac{r^2}{2}\Big)} \cdot \frac{1}{1-\tanh(\mu+\sigma r)^2}\cdot \frac{1}{\epsilon}.
\end{aligned}
\end{equation*}
By calculation, we have
\begin{equation*}
\begin{aligned}
    p(\delta) = & \frac{1}{4\sqrt{2\pi}\sigma\epsilon}\Big[\exp{\Big(-\frac{r^2}{2}+2\sigma r + 2\mu\Big)} + 2\exp{\Big(-\frac{r^2}{2}\Big)} + \exp{\Big(-\frac{r^2}{2}-2\sigma r - 2\mu\Big)}\Big] \\ \leq & \frac{1}{4\sqrt{2\pi}\sigma\epsilon}\Big[\exp{(2\sigma^2 + 2\mu)} + 2 + \exp{(2\sigma^2 - 2\mu)}\Big] \\ \leq & \frac{1}{4\sqrt{2\pi}\kappa_{\sigma}^{lo}\epsilon}\Big[2\exp{(2(\kappa_{\sigma}^{up})^2 + 2\kappa_{\mu})} + 2\Big].
\end{aligned}
\end{equation*}
Hence, $p(\delta)$ is bounded. And the probability density function $p_{\vect{\phi}_i}(\vect{\delta}_i)$ is also bounded since it equals to the product of $p(\delta)$ across all dimensions.

We next prove $p(\delta)$ is Lipschitz continuous at $\delta\in(-\epsilon,\epsilon)$. By calculating the derivative of $p(\delta)$, we have
\begin{equation*}
\begin{aligned}
    p'(\delta) = & \frac{1}{\sqrt{2\pi}\sigma}\exp{\Big(-\frac{(\frac{1}{2}\log\frac{\epsilon+\delta}{\epsilon-\delta}-\mu)^2}{2\sigma^2}\Big)} \cdot \Big[\frac{2\epsilon\delta}{(\epsilon^2-\delta^2)^2} + \frac{\frac{1}{2}\log\frac{\epsilon+\delta}{\epsilon-\delta}-\mu}{\sigma^2} \cdot \big(\frac{\epsilon}{\epsilon^2-\delta^2}\big)^2\Big] \\ = & \frac{1}{\sqrt{2\pi}\sigma}\exp{\Big(-\frac{r^2}{2}\Big)} \cdot \Big[\frac{2\tanh(\mu+\sigma r)}{\epsilon^2(1-\tanh(\mu+\sigma r)^2)^2} + \frac{r}{\sigma\epsilon^2(1-\tanh(\mu+\sigma r)^2)^2}\Big].
\end{aligned}
\end{equation*}
Note that although $p'(\delta)$ has a more complicated form, the quadratic term inside $\exp$ is still $-\frac{r^2}{2}$. Hence, $p'(\delta)$ can also be bounded by a constant. Then $p(\delta)$ as well as $p_{\vect{\phi}_i}(\vect{\delta}_i)$ are Lipschitz continuous. The Lipschitz constant only concerns with $\epsilon$, $\kappa_{\mu}$, $\kappa_{\sigma}^{lo}$, and $\kappa_{\sigma}^{up}$. Hence, the set of explicit distributions in $\mathcal{P}$ with a common Lipschitz constant is equicontinuous.

Combining the results, we prove that the probability density functions of the set of explicit adversarial distributions defined in Eq.~\eqref{eq:explicit} are bound and equicontinuous, which satisfies Assumption~\ref{ass:2}.
\end{proof}

\begin{remark}
Assumption~\ref{ass:2} is used to make the search space $\mathcal{P}$ of the inner problem in ADT compact, as can be seen in Lemma~\ref{lemma:1}. However, it is a sufficient bot not necessary condition of making $\mathcal{P}$ compact. For example, if $\mathcal{P}$ only contains Delta distributions, ADT degenerates to the AT formulation in Eq.~\eqref{eq:at} and $\mathcal{P}$ can be represented by $\mathcal{S}$. In this case, it is easy to see that Assumption~\ref{ass:2} is not satisfied but the search space of the inner problem is also compact.
\end{remark}

\subsection{Proof of Eq.~\eqref{eq:11}}\label{sec:b-2}
%The variable $\vect{\delta}_i$ has the following sampling process
%\begin{equation*}
%\label{eq:explicit}
%    \vect{\delta}_i = \epsilon\cdot\tanh (\vect{u}_i), \quad \vect{u}_i \sim \mathcal{N}(\vect{\mu}_i, \mathrm{diag}(\vect{\sigma}_i^2)),
%\end{equation*}
%whose negative log density is
%\begin{equation*}
%    \sum_{j=1}^d\big(\frac{1}{2}(\vect{r}^{(j)})^2+\frac{\log2\pi}{2}+\log\vect{\sigma}_i^{(j)}+\log(1-\tanh(\vect{\mu}_i^{(j)} + \vect{\sigma}_i^{(j)}\vect{r}^{(j)})^2)+\log\epsilon\big),
%\end{equation*}
%where the superscript $j$ denotes the $j$-th element of a vector.
\begin{proof}
Due to the usage of the diagonal covariance matrix, each dimension in the sampled perturbation $\vect{\delta}_i$ is independent. Thus we can simply calculate the negative log density in each dimension of $\vect{\delta}_i$. For clarity, we also denote $\vect{\mu}_i^{(j)}$, $\vect{\sigma}_i^{(j)}$, $\vect{r}^{(j)}$, $\vect{u}_i^{(j)}$, and $\vect{\delta}_i^{(j)}$ as $\mu$, $\sigma$, $r$, $u$, and $\delta$, respectively. Based on the sampling procedure in Eq.~\eqref{eq:explicit}, we have $\delta=\epsilon \cdot\tanh(u)$ and $u=\mu+\sigma r$. 

Note that $r$ has density: $p(r) = \frac{1}{\sqrt{2\pi}}\exp{(-\frac{r^2}{2})}$. Apply the \emph{transformation of variable} approach, we have the density of $u$ as
\begin{equation*}
    p(u) =  \frac{1}{\sqrt{2\pi}}\exp{(-\frac{r^2}{2})} \cdot \big|\frac{d}{du}(\frac{u-\mu}{\sigma})\big|  =  \frac{1}{\sqrt{2\pi}\sigma}\exp{(-\frac{r^2}{2})}.
\end{equation*}
Let $\beta=\tanh(u)$, then the inverse transformation is $u = \tanh^{-1}(\beta)=\frac{1}{2}\log(\frac{1+\beta}{1-\beta})$, whose derivative w.r.t. $\beta$ is $\frac{1}{1-\beta^2}$. 

Then, by applying the \emph{transformation of variable} approach again, we have the density of $\beta$ as
\begin{equation*}
    p(\beta) =  \frac{1}{\sqrt{2\pi}\sigma}\exp{(-\frac{r^2}{2})} \cdot \frac{1}{1-\beta^2} =  \frac{1}{\sqrt{2\pi}\sigma}\exp{(-\frac{r^2}{2})} \cdot \frac{1}{1-\tanh(\mu+\sigma r)^2}.
\end{equation*}
Therefore, the density of $\delta$ which equals to $\epsilon\cdot\beta$ can be derived similarly, and eventually we obtain
\begin{equation*}
\begin{gathered}
    p(\delta) = \frac{1}{\sqrt{2\pi}\sigma}\exp{(-\frac{r^2}{2})} \cdot \frac{1}{1-\tanh(\mu+\sigma r)^2}\cdot \frac{1}{\epsilon}.
\end{gathered}
\end{equation*}
Consequently, the negative log density of $p(\delta)$ is
\begin{equation*}
    - \log p(\delta) = \frac{r^2}{2}+\frac{\log 2\pi}{2} + \log \sigma + \log(1-\tanh(\mu+\sigma r)^2) + \log\epsilon.
\end{equation*}

Sum over all of the dimensions and we complete the proof of Eq.~\eqref{eq:11}.
\end{proof}

\subsection{Proof of Eq.~\eqref{eq:ent-var}}\label{sec:b-3}
%Given an example $\vect{x}_i$, we define an implicit adversarial distribution $p_{\vect{\phi}}(\vect{\delta}_i|\vect{x}_i)$ in the form of $\vect{\delta}_i=g_{\vect{\phi}}(\vect{z};\vect{x}_i),\vect{z} \sim p(\vect{z})$, where $g_{\vect{\phi}}$ denotes a $\vect{\phi}$-parameterized generator network. Then we can maximize the following variational lower bound to maximize the entropy of $p_{\vect{\phi}}(\vect{\delta}_i|\vect{x}_i)$, as
%\begin{equation*}
%\begin{split}
%    \mathcal{H}(p_{\vect{\phi}}(\vect{\delta}_i|\vect{x}_i)) \geq \mathcal{U}(q) = \mathbb{E}_{p(\vect{z})}\log q(\vect{z} | g_{\vect{\phi}}(\vect{z};\vect{x}_i)) + c
%\end{split}
%\end{equation*}
%where $c$ is a constant and $q(\cdot|\cdot)$ is an introduced variational distribution.
\begin{proof}
We mainly follow~\cite{dai2017calibrating} to provide the proof. Typically, we can view the Dirac generation distribution $p_{\vect{\phi}}(\vect{\delta}_i|\vect{x}_i, \vect{z})$ as a peaked Gaussian with a fixed, diagonal covariance, then it will have a constant entropy.
Considering $\vect{x}_i$ as a given condition, we can simply rewrite the generation distribution as $p_{\vect{\phi},i}(\vect{\delta}_i|\vect{z})$. Then
we can define the joint distribution over $\vect{\delta}_i$ and $\vect{z}$ as $p_{\vect{\phi},i}(\vect{\delta}_i,\vect{z})=p_{\vect{\phi},i}(\vect{\delta}_i|\vect{z})p_{\vect{\phi},i}(\vect{z})$. $p_{\vect{\phi},i}(\vect{z})=p(\vect{z})$ is simply a predefined prior with a constant entropy. Then, we can further define the marginal $p_{\vect{\phi},i}(\vect{\delta}_i)$ whose entropy is of our interest and the posterior $p_{\vect{\phi},i}(\vect{z}|\vect{\delta}_i)$. Consider the mutual information between $\vect{\delta}_i$ and $\vect{z}$
\begin{equation*}
        \mathcal{I}(p_{\vect{\phi},i}(\vect{\delta}_i);p_{\vect{\phi},i}(\vect{z}))=  \mathcal{H}(p_{\vect{\phi},i}(\vect{\delta}_i)) - \mathcal{H}(p_{\vect{\phi},i}(\vect{\delta}_i|\vect{z}))
        =  \mathcal{H}(p_{\vect{\phi},i}(\vect{z})) - \mathcal{H}(p_{\vect{\phi},i}(\vect{z}|\vect{\delta}_i)).
\end{equation*}
Thus, we can calculate the entropy of $\vect{\delta}_i$ as
\begin{equation*}
    \begin{aligned}
    \mathcal{H}(p_{\vect{\phi},i}(\vect{\delta}_i)) =  \mathcal{H}(p_{\vect{\phi},i}(\vect{z})) - \mathcal{H}(p_{\vect{\phi},i}(\vect{z}|\vect{\delta}_i)) + \mathcal{H}(p_{\vect{\phi},i}(\vect{\delta}_i|\vect{z})).
\end{aligned}
\end{equation*}
As stated, the first term and the last term are constant w.r.t. the parameter $\vect{\phi}$. Therefore, maximizing $\mathcal{H}(p_{\vect{\phi},i}(\vect{\delta}_i))$ corresponds to maximizing the negative conditional entropy
\begin{equation*}
    \begin{gathered}
    -\mathcal{H}(p_{\vect{\phi},i}(\vect{z}|\vect{\delta}_i))=\mathbb{E}_{\vect{\delta}_i \sim p_{\vect{\phi},i}(\vect{\delta}_i)}\big[\mathbb{E}_{\vect{z}\sim p_{\vect{\phi},i}(\vect{z}|\vect{\delta}_i)}[\log p_{\vect{\phi},i}(\vect{z}|\vect{\delta}_i)]\big].
\end{gathered}
\end{equation*}
We still cannot optimize this as we have no access to the posterior. As an alternative, we resort to the variational inference technique to tackle this problem. We introduce a variational distribution $q(\vect{z}|\vect{\delta}_i)$ to approximate the true posterior, and derive the following lower bound
\begin{equation*}
    \begin{aligned}
    -\mathcal{H}(p_{\vect{\phi},i}(\vect{z}|\vect{\delta}_i))&=\mathbb{E}_{\vect{\delta}_i \sim p_{\vect{\phi},i}(\vect{\delta}_i)}\big[\mathbb{E}_{\vect{z}\sim p_{\vect{\phi},i}(\vect{z}|\vect{\delta}_i)}[\log q(\vect{z}|\vect{\delta}_i)]\big] + \mathcal{D}_{KL} (p_{\vect{\phi},i}(\vect{z}|\vect{\delta}_i)||q(\vect{z}|\vect{\delta}_i)) \\  &\geq \mathbb{E}_{\vect{\delta}_i \sim p_{\vect{\phi},i}(\vect{\delta}_i)}\big[\mathbb{E}_{\vect{z}\sim p_{\vect{\phi},i}(\vect{z}|\vect{\delta}_i)}[\log q(\vect{z}|\vect{\delta}_i)]\big]
    \\ &= \mathbb{E}_{\vect{z}, \vect{\delta}_i \sim p_{\vect{\phi},i}(\vect{z},\vect{\delta}_i)}[\log q(\vect{z}|\vect{\delta}_i)]
    \\ &= \underbrace{\mathbb{E}_{\vect{z} \sim p_{\vect{\phi},i}(\vect{z})}\big[\mathbb{E}_{\vect{\delta}_i\sim p_{\vect{\phi},i}(\vect{\delta}_i|\vect{z})}[\log q(\vect{z}|\vect{\delta}_i)]\big]}_{\mathcal{U}'(q)},
\end{aligned}
\end{equation*}
where $\mathcal{D}_{KL}$ represents the Kullback–Leibler divergence between distributions.
Note that $p_{\vect{\phi},i}(\vect{z})=p(\vect{z})$ is a prior and $p_{\vect{\phi},i}(\vect{\delta}_i|\vect{z})=p_{\vect{\phi}}(\vect{\delta}_i|\vect{x}_i, \vect{z})$ is Dirac distribution located at $\vect{\delta}_i=g_{\vect{\phi}}(\vect{z};\vect{x}_i)$. Thus, we can write the lower bound of the entropy  $\mathcal{U}(q)$ as
\begin{equation*}
    \mathcal{H}(p_{\vect{\phi}}(\vect{\delta}_i|\vect{x}_i)) \geq \mathcal{U}(q) = \mathcal{U}'(q) + c =  \mathbb{E}_{\vect{z} \sim p(\vect{z})}\log q(\vect{z}|g_{\vect{\phi}}(\vect{z};\vect{x}_i)) + c,
\end{equation*}
which can be optimized effectively via Monte Carlo integration and standard back-propagation. Then we finish the proof of Eq.~\eqref{eq:ent-var}.
\end{proof}

\section{Detailed experimental settings} \label{sec:c}
We provide the detailed experimental settings in this section.
All of the experiments are conducted on NVIDIA 2080 Ti GPUs.
%The source code of ADT is available at \url{https://github.com/dongyp13/Adversarial-Distributional-Training}.

\subsection{Datasets} 
We choose the CIFAR-10~\cite{krizhevsky2009learning}, CIFAR-100~\cite{krizhevsky2009learning}, and SVHN~\cite{netzer2011reading} datasets to conduct the experiments.
CIFAR consists of a training set of $50,000$ and a test set of $10,000$ color images of resolution $32\times 32$ with $10$ classes in CIFAR-10 and $100$ classes in CIFAR-100.
SVHN is a $10$-class house number classification dataset with $73,257$ training images and $26,032$ test images.
During training, we perform standard data augmentation (i.e., horizontal flips and random crops from images with $4$ pixels padded on each side) on CIFAR-10 and CIFAR-100, and use no data augmentation on SVHN. 
We do not use any data augmentation during testing.

\subsection{Network architectures}
For the generator network in ADT\textsubscript{EXP-AM} and ADT\textsubscript{IMP-AM}, we adopt a popular image-to-image architecture which has shown promise in neural style transfer and super-resolution~\cite{johnson2016perceptual,zhu2017unpaired}. The network contains $3$ residual blocks~\cite{he2016deep}, with two extra convolutions at the beginning and the end. All convolutions in the generator have stride $1$, and are immediately followed by batch normalization~\cite{ioffe2015batch} and ReLU activation. 

As found by~\cite{chen2018learning}, taking only the natural images as inputs to the generator network can lead to poor results. And they suggest to input the classifier's gradients as well. Based on this finding, we calculate the gradient of the loss function at the natural input $\vect{g}_i^1 = \nabla_{\vect{x}}\mathcal{L}(f_{\vect{\theta}}(\vect{x}_i),y_i)$, as well as the gradient of the loss function at the FGSM adversarial example $\vect{g}_i^2 = \nabla_{\vect{x}}\mathcal{L}(f_{\vect{\theta}}(\vect{x}_i + \vect{\delta}_i^{\mathrm{FGSM}}),y_i)$, where $\vect{\delta}_i^{\mathrm{FGSM}} = \epsilon\cdot\mathrm{sign}(\nabla_{\vect{x}}\mathcal{L}(f_{\vect{\theta}}(\vect{x}_i),y_i))$, and then input $[\vect{x}_i, \vect{g}_i^1, \vect{g}_i^2]$ to the generator network.

In ADT\textsubscript{EXP-AM}, the generator has $6$ output channels to deliver the parameters (i.e., mean and standard derivation) of the explicit adversarial distributions.
In ADT\textsubscript{IMP-AM}, for each input we sample a $64$-dim i.i.d. $\vect{z}$ from a uniform distribution $\mathrm{U}(-1,1)$, which is encoded with $2$ fully connected (FC) layers and then fed into the generator along with the input image and gradients.

We elaborate the architectures of the generator networks in Table~\ref{table:archs}, and the architecture of $q$ in ADT\textsubscript{IMP-AM} in Table~\ref{table:arch-q}. In these tables, ``C$\times$H$\times$W'' means a convolutional layer with C filters size H$\times$W, which is followed by batch normalization~\cite{ioffe2015batch} and a ReLU nonlinearity (or LeakyReLU for layers in Table~\ref{table:arch-q}), except the last layers in the architectures. We use the residual block design in~\cite{he2016deep}, which is composed of two $3\times3$ convolutions and a residual connection.

\begin{table}[t]
  \caption{The network architectures used for the generators.}
  \centering
  \begin{tabular}{c||c}
  \hline
In ADT\textsubscript{EXP-AM} & In ADT\textsubscript{IMP-AM}\\
\hline
    input & $\vect{z}$ \\
    $256\times3\times3$ conv& $256$-dim fc layer\\
    Residual block, $512$ filters& $1024$-dim fc layer\\
    Residual block, $512$ filters& reshape to $1\times32\times32$\\
    Residual block, $512$ filters& concat with input\\
    $6\times3\times3$ conv & $256\times3\times3$ conv \\
     & Residual block, $512$ filters \\
    & Residual block, $512$ filters \\
    & Residual block, $512$ filters \\
     & $3\times3\times3$ conv \\
  \hline
   \end{tabular}
  \label{table:archs}
\end{table}

\begin{table}[t]
  \caption{The network architecture used for instantiating the variational distribution $q$ in ADT\textsubscript{IMP-AM}.}
  \centering
  \begin{tabular}{c}
  \hline
Layers\\
\hline
    input\\
    $32\times5\times5$, stride $1$\\
    $64\times4\times4$, stride $2$\\
    $128\times4\times4$, stride $1$\\
    $256\times4\times4$, stride $2$\\
    Global average pooling\\
    $128\times1\times1$, stride $1$\\
    
  \hline
   \end{tabular}
  \label{table:arch-q}
\end{table}

\subsection{Training details}
The classifier is trained using SGD with momentum $0.9$, weight decay $2\times 10^{-4}$, and batch size $64$. The initial learning rate is $0.1$, which is reduced to $0.01$ in the $75$-th epoch. We stop training after $76$ epochs.
For ADT\textsubscript{EXP}, we adopt Adam~\cite{Kingma2014} for optimizing the distribution parameters $\vect{\phi}_i$. We set the learning rate for $\vect{\phi}_i$ as $0.3$, the momentum as $(0.0, 0.0)$, the number of optimization steps as $T=7$, and the number of MC samples to estimate the gradient in each step as $k=5$.
For ADT\textsubscript{EXP-AM} and  ADT\textsubscript{IMP-AM}, we use only one MC sample for gradient estimation and use Adam with momentum $(0.5, 0.999)$ and learning rate $2\times 10^{-4}$ to optimize the parameter $\vect{\phi}$ of the generator network. We also adopt Adam with learning rate $2\times 10^{-4}$ to optimize the parameter $\vect{\psi}$ of the introduced variational in ADT\textsubscript{IMP-AM}.

\subsection{Baselines}
Our primary baselines include: 1) standard training on the clean images (\textbf{\emph{Standard}}); 2) adversarial training on the PGD adversarial examples (\textbf{\emph{AT\textsubscript{PGD}}})~\cite{madry2017towards}. 
Standard and AT\textsubscript{PGD} are trained with the same configurations specified above. 
For training AT\textsubscript{PGD}, we perform PGD with $T=7$ steps, and step size $\alpha=\nicefrac{\epsilon}{4}$, which are the same as in~\cite{madry2017towards}.
On CIFAR-10, we incorporate several additional baselines, including: 1) the pretrained AT\textsubscript{PGD} model (\textbf{\emph{AT\textsubscript{PGD}$^{\dagger}$}}) released by~\cite{madry2017towards}; 2) adversarial training on the targeted FGSM adversarial examples (\textbf{\emph{AT\textsubscript{FGSM}}})~\cite{kurakin2016adversarial}; 3) adversarial logit pairing (\textbf{\emph{ALP}})~\cite{kannan2018adversarial}; and 4) feature scattering-based adversarial training (\textbf{\emph{FeaScatter}})~\cite{zhang2019defense}.
We implement AT\textsubscript{FGSM} and ALP by ourselves using the same training configuration specified above and use the pretrained model of FeaScatter.
Note that all of these models have the same network architecture for a fair comparison.

\subsection{A feature attack for white-box evaluation}\label{sec:c-5}
We incorporate a feature attack (FeaAttack)~\cite{lin2019feature} for white-box robustness evaluation in this paper.
The algorithm of FeaAttack is introduced below.
Given a natural input $\vect{x}$, FeaAttack first finds a target image $\vect{x}'$ belonging to a different class. It minimizes the cosine similarity between the feature representations of the adversarial example and $\vect{x}'$ as
\begin{equation*}
    \vect{\delta}^* = \argmin_{\vect{\delta}\in\mathcal{S}} \mathcal{L}_{cos} (f'_{\vect{\theta}}(\vect{x} + \vect{\delta}), f'_{\vect{\theta}}(\vect{x}')),
\end{equation*}
where $f'_{\vect{\theta}}(\cdot)$ returns the feature representation before the global average pooling layer for an input, and $\mathcal{L}_{cos}$ is the cosine similarity between two features.
FeaAttack solves this objective function by
\begin{equation*}
    \vect{\delta}^{t+1}=\Pi_{\mathcal{S}}\big(\vect{\delta}^t - \alpha\cdot\mathrm{sign}(\nabla_{\vect{x}}\mathcal{L}_{cos}(f'_{\vect{\theta}}(\vect{x} + \vect{\delta}^t), f'_{\vect{\theta}}(\vect{x}')))\big).
\end{equation*}
$\vect{\delta}^0$ is initialized uniformly in $\mathcal{S}$.
In our experiments, we set $\alpha=\nicefrac{\epsilon}{8}$ and the number of optimization steps as $50$. For each natural input, we randomly select $200$ target images to conduct $200$ attacks, and report a successful attack when one of them can cause misclassification of the model.

\begin{table}[t]
\caption{Classification accuracy of the three proposed methods and baselines on CIFAR-100 and SVHN under white-box attacks. We mark the best results for each attack and the overall results that outperform the baselines in \textbf{bold}, and the overall best result in \textcolor{blue}{\textbf{blue}}.}
  \centering
 \small
  %\begin{sc}
  \begin{tabular}{c||p{6.8ex}<{\centering}|p{6.8ex}<{\centering}|p{7.5ex}<{\centering}|p{9ex}<{\centering}|p{6.8ex}<{\centering}|p{6.8ex}<{\centering}|p{9ex}<{\centering}|p{6.8ex}<{\centering}}
  \hline
Model & $\mathcal{A}_{\mathrm{nat}}$ & FGSM & PGD-20 & PGD-100 & MIM & C\&W & FeaAttack &  $\mathcal{A}_{\mathrm{rob}}$ \\
\hline\hline
\multicolumn{9}{c}{CIFAR-100, $\epsilon=8/255$}\\
\hline
Standard & \bf78.59\% & 8.73\% & 0.02\% & 0.01\% & 0.02\% & 0.00\% & 0.00\% & 0.00\%\\
AT\textsubscript{PGD} & 61.45\% & 30.78\% & 25.71\% & 25.40\% & 26.60\% & 25.80\% & 33.95\% & 24.49\%\\
\hline
ADT\textsubscript{EXP} & 62.70\% & 34.22\% & 28.96\% & \bf28.60\% & \bf29.83\% & \bf28.99\% & \bf35.07\% & \textcolor{blue}{\bf27.13\%}\\
ADT\textsubscript{EXP-AM} & 62.84\% & 36.28\% & 29.01\% & 28.46\% & 29.68\% & 28.78\% & 34.91\% & \bf26.87\%\\
ADT\textsubscript{IMP-AM} & 64.07\% & \bf39.39\% & \bf29.40\% & 28.43\% & 29.64\% & 28.76\% & 35.00\% & \bf26.80\%\\
  \hline\hline
 \multicolumn{9}{c}{SVHN, $\epsilon=4/255$}\\
\hline
Standard & \bf96.12\% & 39.05\% & 3.64\% & 2.95\% & 4.08\% & 3.91\% & 2.14\% & 2.14\%\\
AT\textsubscript{PGD} & 95.07\% & 82.19\% & 74.22\% & 73.79\% & 74.56\% & 74.77\% & 73.51\% & 73.38\%\\
\hline
ADT\textsubscript{EXP} & 95.70\% & 86.72\% & \bf77.01\% & \bf76.62\% & \bf77.18\% & \bf77.50\% & \bf75.64\% & \textcolor{blue}{\bf 75.55\%}\\
ADT\textsubscript{EXP-AM} & 95.67\% & 85.24\% & 76.12\% & 75.58\% & 76.63\% & 76.70\% & 75.20\% & \bf75.00\%\\
ADT\textsubscript{IMP-AM} & 95.62\% & \bf86.73\% & 75.61\% & 74.85\% & 75.91\% & 76.12\% & 74.24\% & \bf74.13\%\\
\hline
   \end{tabular}
  \label{table:other2}
\end{table}

\section{Supplementary experimental results}
We provide more experimental results in this section.

\subsection{Full results on CIFAR-100 and SVHN}\label{sec:d-1}
We provide the full experimental results of Standard, AT\textsubscript{PGD}, ADT\textsubscript{EXP}, ADT\textsubscript{EXP-AM}, and ADT\textsubscript{IMP-AM} under all adopted white-box attacks on CIFAR-100 and SVHN in Table~\ref{table:other2}.

\begin{table}[t]
\caption{Classification accuracy of TRADES and the three ADT-based methods trained with the TRADES loss on CIFAR-10 under white-box attacks with $\epsilon=8/255$. We mark the best results for each attack and the overall results that outperform the baselines in \textbf{bold}, and the overall best result in \textcolor{blue}{\textbf{blue}}.}
  \centering
 \small
 \setlength{\tabcolsep}{5pt}
  %\begin{sc}
  \begin{tabular}{c|c|p{6.8ex}<{\centering}|p{6.8ex}<{\centering}|p{7.5ex}<{\centering}|p{9ex}<{\centering}|p{6.8ex}<{\centering}|p{6.8ex}<{\centering}|p{9ex}<{\centering}|p{6.8ex}<{\centering}}
  \hline
Model & $\beta$ & $\mathcal{A}_{\mathrm{nat}}$ & FGSM & PGD-20 & PGD-100 & MIM & C\&W & FeaAttack &  $\mathcal{A}_{\mathrm{rob}}$ \\
\hline\hline
TRADES & 1.0 & 87.99\% & 57.67\% & 51.08\% & 48.41\% & 53.32\% & 49.29\% & 51.07\% & 47.75\%\\
ADT\textsubscript{EXP} & 1.0 & \bf89.74\% & 59.47\% & 52.39\% & 49.88\% & 54.74\% & 50.75\% & 51.29\% & \bf49.05\%\\
ADT\textsubscript{EXP-AM} & 1.0 & 88.86\% & 62.89\% & \bf54.44\% & \bf51.66\% & \bf56.09\% & \bf52.33\% & \bf54.61\% & \textcolor{blue}{\bf50.78\%}\\
ADT\textsubscript{IMP-AM} & 1.0 & 88.80\% & \bf68.35\% & 54.22\% & 51.09\% & 54.95\% & 51.84\% & 54.19\% & \bf50.14\%\\
  \hline\hline
TRADES & 6.0 & 84.02\% & 60.08\% & 56.06\% & 54.49\% & 57.27\% & 53.62\% & 55.18\% & 52.64\%\\
ADT\textsubscript{EXP} & 6.0 & 84.66\% & 61.72\% & 57.71\% & \bf56.17\% & \bf58.74\% & \bf55.16\% & 56.65\% & \textcolor{blue}{\bf 54.21\%}\\
ADT\textsubscript{EXP-AM} & 6.0 & 84.85\% & 66.09\% & 57.67\% & 55.73\% & 58.38\% & 54.79\% & 58.94\% & \bf54.09\%\\
ADT\textsubscript{IMP-AM} & 6.0 & \bf84.96\% & \bf68.34\% & 
\bf57.82\% & 55.45\% & 58.58\% & 54.36\% & \bf59.01\% & \bf53.66\%\\
\hline
   \end{tabular}
  \label{table:trades2}
\end{table}

\subsection{Full results on TRADES}\label{sec:d-2}
In TRADES~\cite{zhang2019theoretically}, the minimax optimization problem is formulated as
\begin{equation*}
    \min_{\vect{\theta}}\frac{1}{n}\sum_{i=1}^{n}\Big\{\mathcal{L}(f_{\vect{\theta}}(\vect{x}_i), y_i)+\beta\cdot\max_{\vect{\delta}_i\in\mathcal{S}}\mathcal{D}_{\mathrm{KL}}(f_{\vect{\theta}}(\vect{x}_i + \vect{\delta}_i), f_{\vect{\theta}}(\vect{x}_i))\Big\},
\end{equation*}
where $\beta$ is a hyperparameter balancing the trade-off between natural and robust accuracy. 
The full experimental results of TRADES and the three variants of ADT when integrated with the TRADES loss are shown in Table~\ref{table:trades2}. We evaluate their performance by all adopted white-box attacks and report the worst-case robustness as in Eq.~\eqref{eq:12}.

\subsection{Convergence of learning the explicit adversarial distributions}\label{sec:d-3}

\begin{figure}[t]
\centering
\begin{minipage}{.4\columnwidth}
\centering
\centerline{\includegraphics[width=1.0\linewidth]{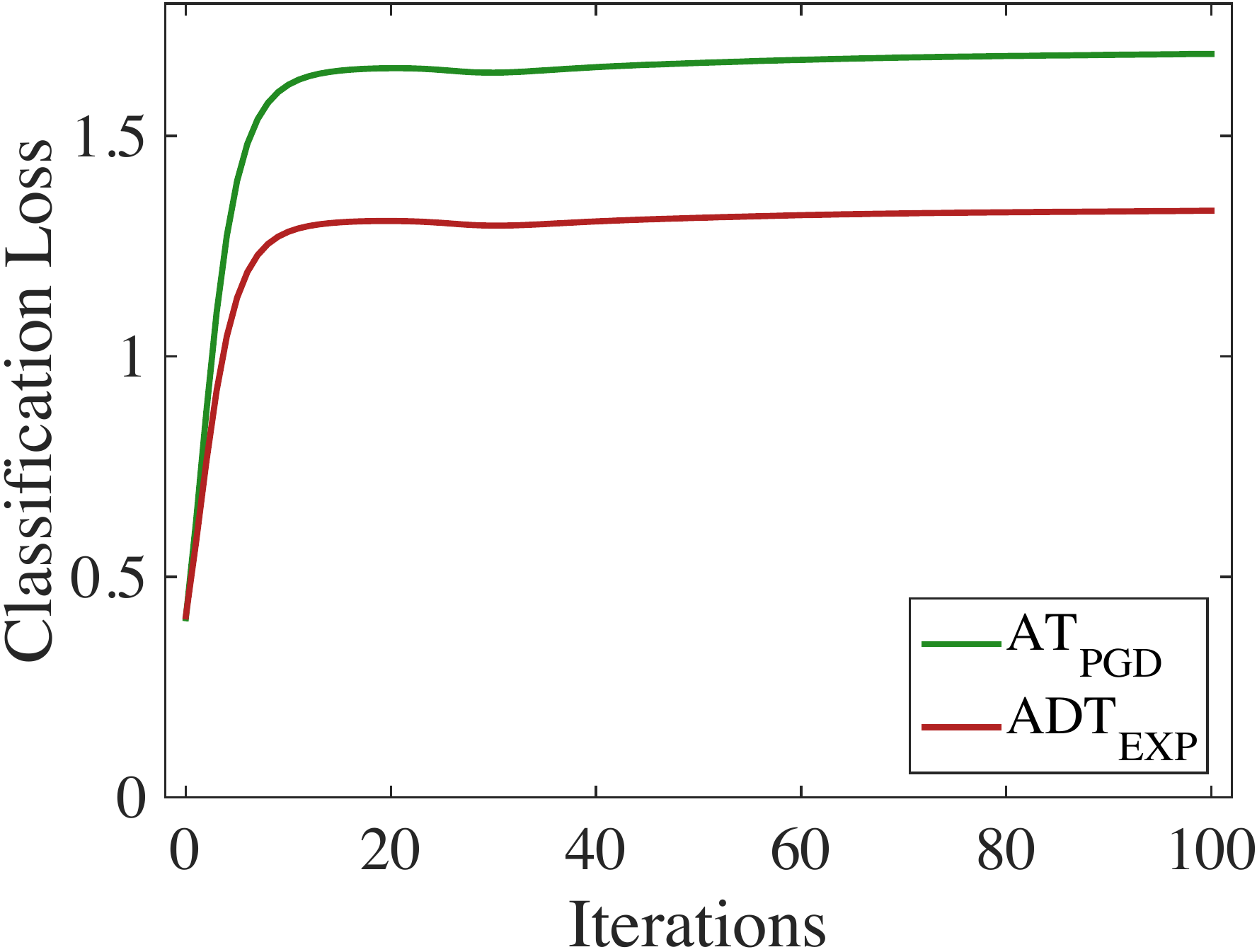}}
\end{minipage}
\hspace{5ex}
\begin{minipage}{.4\columnwidth}
\centering
\centerline{\includegraphics[width=1.0\linewidth]{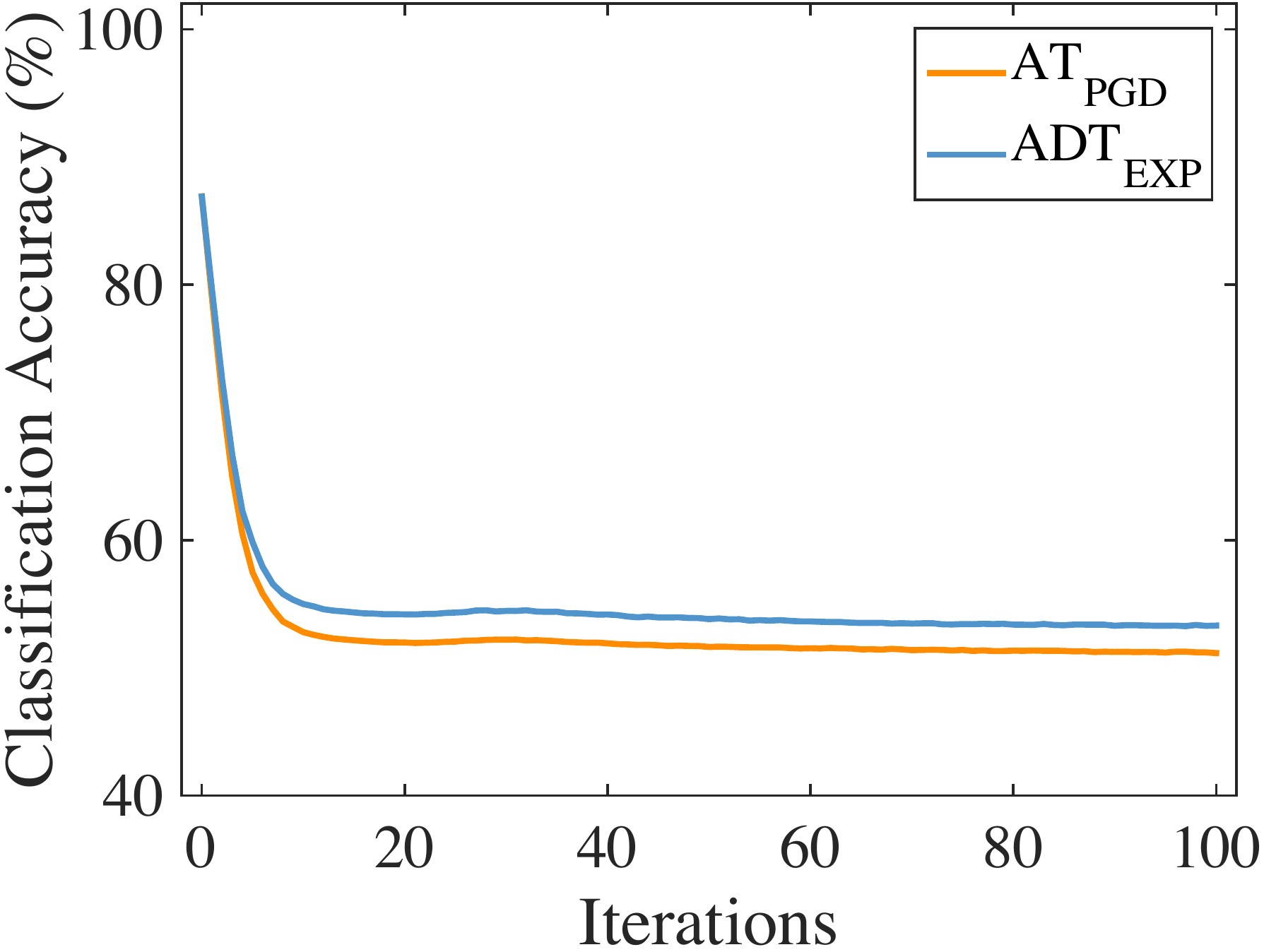}}
\end{minipage}
\caption{Classification loss (i.e., cross-entropy loss) and accuracy (\%) of AT\textsubscript{PGD} and ADT\textsubscript{EXP} under the explicit adversarial distributions attack with different attack iterations.}
\label{fig:convergence}
\end{figure}

We study the convergence of the explicit adversarial distributions introduced in Sec.~\ref{sec:adt_gau} by attacking AT\textsubscript{PGD} and ADT\textsubscript{EXP} with varying iterations. We set the learning rate of $\vect{\phi}_i$ as $0.3$, the momentum as $(0.0,0.0)$, the number of MC samples to estimate the gradient in each step as $k=10$, and vary the attack iterations from $0$ to $100$.
We show the classification loss and accuracy in Fig.~\ref{fig:convergence}. Learning the explicit adversarial distributions can converge soon within a few iterations.

\subsection{Training time}\label{sec:d-4}
\begin{figure}[t]
\centering
\includegraphics[width=0.5\columnwidth]{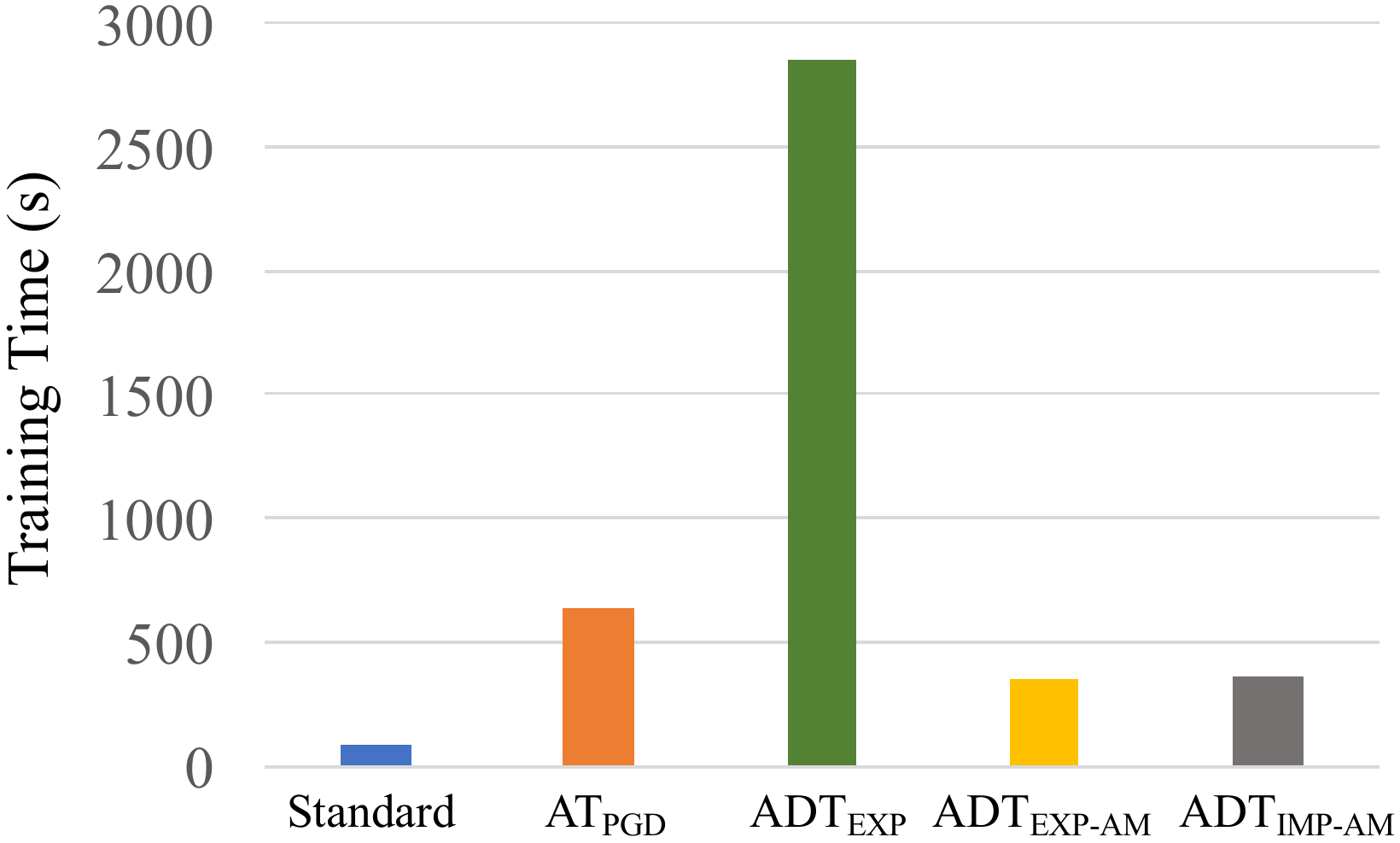}
\caption{The training time (s) for one epoch of Standard, AT\textsubscript{PGD}, ADT\textsubscript{EXP}, ADT\textsubscript{EXP-AM}, and ADT\textsubscript{IMP-AM} on CIFAR-10.}
\label{fig:time}
\end{figure}

We provide the one-epoch training time of Standard, AT\textsubscript{PGD}, ADT\textsubscript{EXP}, ADT\textsubscript{EXP-AM}, and ADT\textsubscript{IMP-AM} on CIFAR-10 in Fig.~\ref{fig:time}.
As can be seen, ADT\textsubscript{EXP} is nearly $5\times$ slower than AT\textsubscript{PGD} since we use $k=5$ MC samples to estimate the gradient w.r.t. the distribution parameters in each step. Nevertheless, by amortizing the adversarial distributions, ADT\textsubscript{EXP-AM} and ADT\textsubscript{IMP-AM} are much faster than ADT\textsubscript{EXP}, and nearly $2\times$ faster than AT\textsubscript{PGD}.

\begin{table}[t]
  \caption{Classification Accuracy of L2L~\cite{chen2018learning}, ADT\textsubscript{EXP-AM}, and ADT\textsubscript{IMP-AM} on CIFAR-10 under white-box attacks with $\epsilon=8/255$.}
  \centering
  \begin{tabular}{c||c|c|c}
  \hline
Model & L2L & ADT\textsubscript{EXP-AM} & ADT\textsubscript{IMP-AM}\\
\hline\hline
 $\mathcal{A}_{\mathrm{nat}}$ & \bf88.15\% & 87.82\% & 88.00\% \\
 \hline
FGSM & \bf65.50\% & 62.42\% & 64.89\%\\
PGD-20 & 48.55\% & 51.95\% & \bf52.28\%\\
PGD-100 & 47.14\% & \bf51.26\% & 51.23\%\\
MIM & 49.03\% & \bf52.99\% & 52.64\%\\
C\&W & 49.22\% & 51.75\% & \bf52.65\%\\
  \hline
   \end{tabular}
  \label{table:l2l}
\end{table}

\subsection{Comparison with \citet{chen2018learning}}\label{sec:d-5}

We further compare ADT with the L2L framework in~\cite{chen2018learning}.
Their method is similar to ours in the sense that they also adopt a generator network to produce adversarial examples, and perform adversarial training on those generated adversarial examples.
The essential different between our methods and theirs is that we propose an adversarial distributional training framework to learn the distributions of adversarial perturbations, while their method is a variant of the vanilla adversarial training with a different approach to solving the inner maximization.

Since the source code is not provided by \citet{chen2018learning}, we tried to reproduce their reported results with the same training configuration specified in their paper, but we failed.
Therefore, we adopt the same configuration as in ADT for training the L2L model.
Table~\ref{table:l2l} shows the results of L2L, ADT\textsubscript{EXP-AM}, and ADT\textsubscript{IMP-AM}, which use the same classifier architecture and generator network. Our ADT-based methods outperform L2L in most cases, showing the advantages of learning the distributions of adversarial perturbations upon finding a single adversarial example.

\end{document}